\documentclass[letterpaper, 10 pt, conference]{URL-ieeeconf}
\IEEEoverridecommandlockouts    

\usepackage{graphics}           
\usepackage{times}              
\usepackage{amsmath}            
\usepackage{amssymb}            
\usepackage{graphicx}
\usepackage{algorithm}
\usepackage[noend]{algpseudocode}
\usepackage{booktabs}
\usepackage{color}
\usepackage{multirow}
\usepackage{subcaption}
\usepackage{rotating}
\usepackage{cite}
\usepackage{pifont}
\usepackage{hyperref}
\definecolor{instructioncolor}{rgb}{.0,.0,.0}
\definecolor{darkred}{rgb}{.804,.196,.196}
\definecolor{darkorange}{rgb}{1.0,.55,.0}
\definecolor{darkgreen}{rgb}{.196,.70,.196}

\newcommand{\cmark}{\textcolor{darkgreen}{\ding{51}}}
\newcommand{\xmark}{\textcolor{darkred}{\ding{55}}}
\newcommand{\trimark}{\textcolor{darkorange}{$\boldsymbol{\triangle}$}}

\newcommand{\changed}[1]{\textcolor{instructioncolor}{#1}}

\def\secref#1{Section~\ref{#1}}
\def\figref#1{Fig.~\ref{#1}}
\def\tabref#1{Table~\ref{#1}}
\def\eqref#1{(\ref{#1})}

\def\vsfigu{\vspace{-0.1cm}}
\def\vsfig{\vspace{-0.3cm}}

\def\vsequ{\vspace{-0.15cm}}
\def\vseq{\vspace{-0.3cm}}

\def\curridx{t}

\newcommand{\rom}[1]{\uppercase\expandafter{\romannumeral #1\relax}}

\makeatletter
\usepackage{xspace}
\DeclareRobustCommand\onedot{\futurelet\@let@token\@onedot}
\def\@onedot{\ifx\@let@token.\else.\null\fi\xspace}
\def\eg{e.g\onedot} 
\def\ie{i.e\onedot}

\def\etal{{\textit{et al}}\onedot}
\makeatother

\def\etalcite#1{\etal~\cite{#1}}

\usepackage{array}
\newcolumntype{L}[1]{>{\raggedright\let\newline\\\arraybackslash\hspace{0pt}}m{#1}}
\newcolumntype{C}[1]{>{\centering\let\newline\\\arraybackslash\hspace{0pt}}m{#1}}
\newcolumntype{R}[1]{>{\raggedleft\let\newline\\\arraybackslash\hspace{0pt}}m{#1}}

\usepackage[table]{xcolor}
\usepackage{soul}
\sethlcolor{lightgray}

\title{\LARGE \bf HeLiMOS: A Dataset for Moving Object Segmentation\\ in 3D Point Clouds From Heterogeneous LiDAR Sensors}

\author{Hyungtae Lim$^{1\dagger}$, Seoyeon Jang$^{1\dagger}$, Benedikt Mersch$^2$, Jens Behley$^2$, Hyun Myung$^{1*}$, and Cyrill Stachniss$^{2,3}$\thanks{$^*$Corresponding author: Hyun Myung (E-mail: \scriptsize{\texttt{hmyung@kaist.ac.kr}})}
  \thanks{$^\dagger$The authors are equally contributed.}
  \thanks{$^{1}$Hyungtae Lim, Seoyeon Jang, and Hyun Myung are with the School of Electrical Engineering, KAIST (Korea Advanced Institute of Science and Technology), Daejeon, Republic of Korea. \hfill \break \indent $^{2}$Benedikt Mersch, Jens Behley, and Cyrill Stachniss are with the Center for Robotics, University of Bonn, Germany \hfill \break
  \indent $^{3}$Cyrill Stachniss is additionally with the Lamarr Institute for Machine Learning and Artificial Intelligence, Germany (E-mail: {\scriptsize{\texttt{cyrill.stachniss@igg.uni-bonn.de}}}).  \hfill \break
  \indent This work was supported by the Technology Innovation Program~(or Industrial Strategic Technology Development Program-Robot Industry Technology Development)~(00427719, Dexterous and Agile Humanoid Robots for Industrial Applications)~funded By the Ministry of Trade Industry~\&~Energy~(MOTIE, Korea), and in part by the National Research Foundation of Korea~(NRF) grant funded by the Korea government~(MSIT)~(No. RS-2024-00348461) and the Basic Science Research Program through the National Research Foundation of Korea~(NRF) funded by the Ministry of Education~(RS-2024-00415018).
  The Korean students are supported by the BK21 FOUR, Republic of Korea.}
}

\begin{document}
\maketitle
\thispagestyle{empty}
\pagestyle{empty}

\begin{abstract}
	Moving object segmentation (MOS) using a 3D light detection and ranging (LiDAR) sensor is crucial for scene understanding and identification of moving objects.
	Despite the availability of various types of 3D LiDAR sensors in the market,
	MOS research still predominantly focuses on 3D point clouds from mechanically spinning omnidirectional LiDAR sensors. Thus, we are, for example, lacking a dataset with MOS labels for point clouds from solid-state LiDAR sensors which have irregular scanning patterns.
	In this paper, we present a labeled dataset, called \textit{HeLiMOS}, that enables to test MOS approaches on four heterogeneous LiDAR sensors, including two solid-state LiDAR sensors.
	Furthermore, we introduce a novel automatic labeling method to substantially reduce the labeling effort required from human annotators.
	To this end, our framework exploits an instance-aware static map building approach and tracking-based false label filtering.
	Finally, we provide experimental results regarding the performance of commonly used state-of-the-art MOS approaches on HeLiMOS that suggest a new direction for a sensor-agnostic MOS, which generally works regardless of the type of LiDAR sensors used to capture 3D point clouds.
	Our dataset is available at \href{https://sites.google.com/view/helimos} {https://sites.google.com/view/helimos}.
\end{abstract}

\section{Introduction}
\label{sec:intro}

Robots need to understand their surroundings, including moving objects, to navigate and act safely. By doing so, robots can avoid collisions, optimize paths, and make informed decisions based on dynamic changes around them.
As one of the solutions, moving object segmentation (MOS) with 3D light detection and ranging~(LiDAR) sensors has been extensively studied, aiming to identify moving objects~\cite{chen2022ral, chen2021ral, sun2022mos3d, kim2022ral-rvmos, mersch2022ral, mersch2023building, wu2024moving, cheng2024mf, gu2023lidar, wang2023insmos}.
By distinguishing between moving objects such as buses, cars, and pedestrians, and static objects such as buildings, walls, and trees,
MOS not only enhances path planning and collision avoidance, but also prevents traces of moving objects, which we call \textit{dynamic points}, from being left in a 3D point cloud map by filtering out these undesirable points at the perception level~\cite{stachniss2005aaai, stachniss2009springerbook, kim2020iros, lim2021ral, lim2023erasor2, zhang2023dynamic}.

Meanwhile, diverse types of 3D LiDAR sensors have been developed, including mechanically spinning, prism-rotating, solid-state, frequency-modulated continuous wave~(FMCW), and flash types.
With the growing need for datasets to evaluate existing approaches across heterogeneous LiDAR sensor setups,
some researchers~\cite{qingqing2022tiers, jung2023helipr} proposed novel large-scale datasets captured by heterogeneous LiDAR sensor setups.
In addition, Mersch~\etalcite{mersch2023building} and Wu~\etalcite{wu2024moving} showed pioneering works by demonstrating the feasibility of MOS with heterogeneous LiDAR configurations.

\begin{figure}[t]
	\centering
	\begin{subfigure}[b]{0.45\textwidth}
		\centering
		\includegraphics[width=1.0\textwidth]{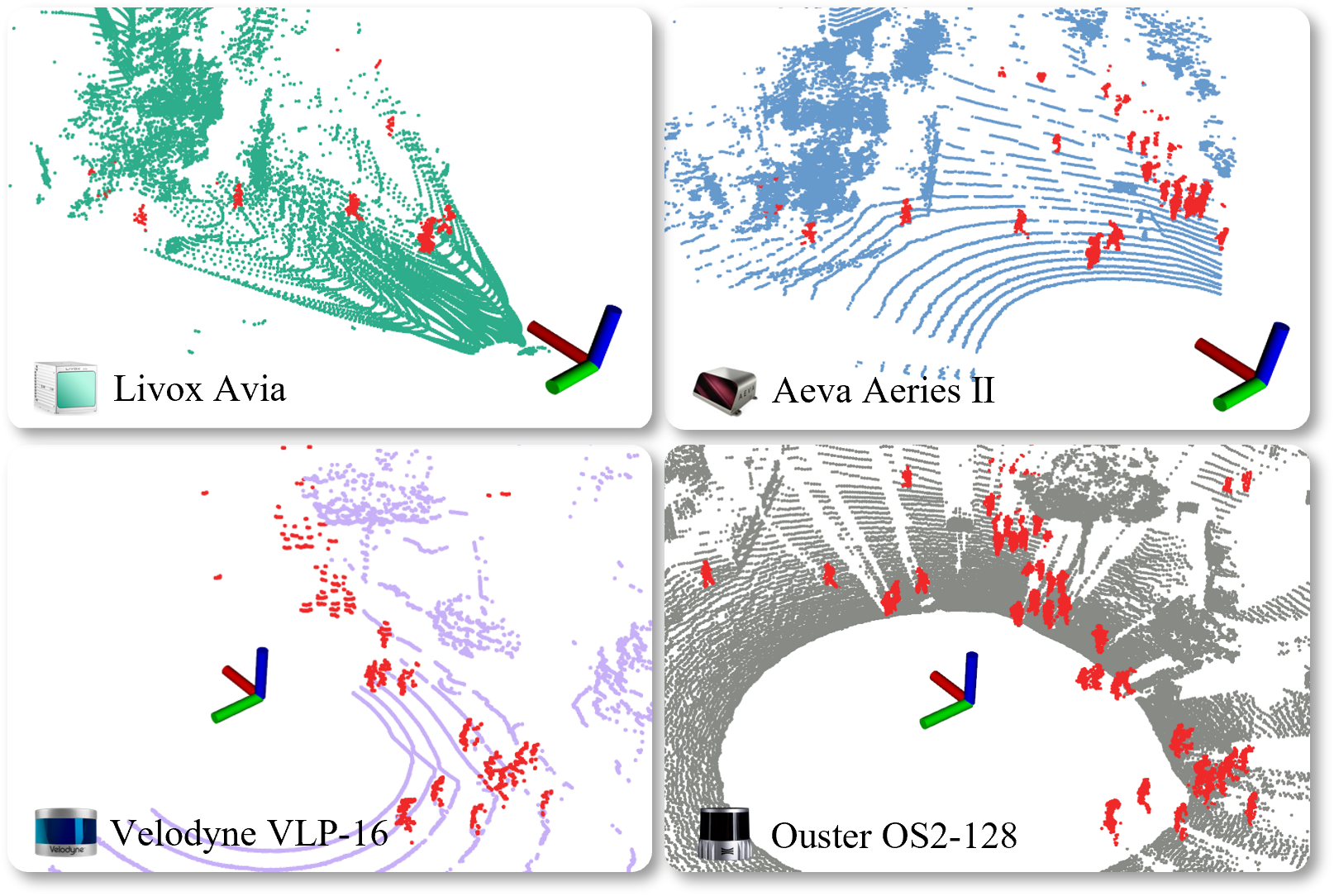}
	\end{subfigure}
	\captionsetup{font=footnotesize}
	\caption{Qualitative examples of our dataset, called \textit{HeLiMOS}. Our dataset provides point-wise moving object segmentation (MOS) annotations for point clouds acquired by heterogeneous 3D LiDAR sensors from the HeLiPR dataset~\cite{jung2023helipr}. Red points indicate the annotated points from moving objects~(best viewed in color).}
	\label{fig:fig1}
	\vsfig
\end{figure}

Despite these efforts, we see that existing public datasets have two limitations for evaluating the generalization capabilities of MOS across heterogeneous LiDAR sensor setups.
First, the aforementioned heterogeneous LiDAR datasets mainly focus on evaluating place recognition~\cite{jung2023helipr} or pose estimation~\cite{qingqing2022tiers} without providing point-wise MOS labels.
Second, while multiple datasets that provide point-wise MOS labels exist~\cite{behley2019iccv, pan2020semanticposs},
these datasets are only acquired by a single omnidirectional LiDAR sensor.
Thus, publicly available datasets with point-wise MOS labels for heterogeneous LiDAR setups are still lacking.

To tackle the insufficiency of MOS labels for heterogeneous LiDAR sensors, as shown in~\figref{fig:fig1},
we build upon the existing HeLiPR dataset~\cite{jung2023helipr} and provide MOS labels that enable the evaluation of MOS across four heterogeneous LiDAR sensor setups, which we call \textit{HeLiMOS}.
Furthermore, sharing the philosophy of the state-of-the-art automatic MOS labeling framework~\cite{chen2022ral},
we propose a novel instance-aware automatic labeling framework to substantially reduce the time needed for manual labeling.
Finally, as a preliminary step, we set up benchmarks for evaluating MOS from an egocentric perspective and static map building from a map-centric perspective.

In summary, our main contributions are threefold:
\begin{itemize}
\item We provide point-wise annotations for a sequence of the HeLiPR dataset, which are captured by real-world multiple heterogeneous LiDAR sensors.
\item We propose an efficient instance-aware automatic labeling framework by employing an instance-aware static map building approach, ERASOR2~\cite{lim2023erasor2}, and tracking-based false label filtering~\cite{jang2023toss}. We also make these MOS labeling tools publicly available.
\item We evaluate state-of-the-art MOS approaches with heterogeneous LiDAR sensor setups as initial benchmarks.
\end{itemize}

We believe this dataset will stimulate further research, suggest new research directions, and enable reliable evaluation of novel algorithms.

\section{Related Work}
\label{sec:related}

Over the past decade, numerous impactful datasets for autonomous vehicles have been released, providing novel benchmarks. One of the renowned datasets is the KITTI dataset~\cite{geiger2012cvpr}, which provides both odometry and various perception benchmarks.
Influenced by the KITTI dataset, existing datasets have evolved in two main directions in terms of (a)~odometry and place recognition and (b)~perception, as presented in~\tabref{table:summary}.

From the viewpoint of odometry and place recognition, the KITTI dataset has few loop closing situations and environmental changes, with only a single omnidirectional LiDAR sensor for a short data collection span.
To provide more challenging environments for odometry and place recognition tasks~\cite{yin2024survey}, Carlevaris~\etalcite{carlevaris2016nclt} and Jeong~\etalcite{jeong2019complex} proposed the NCLT and Complex Urban datasets, respectively, acquired by multiple 2D and 3D omnidirectional LiDAR sensors.
Kim~\etalcite{kim2020mulran} proposed the MulRan dataset focusing on multi-modal long-term mapping and place recognition by employing a 3D LiDAR sensor and an omnidirectional radar sensor.

As a further study, Carballo~\etalcite{carballo2020libre} proposed the LIBRE dataset, which consists of point clouds from ten different omnidirectional LiDAR sensors.
However, all the deployed sensors are omnidirectional LiDAR sensors, implying that all the sensors are homogeneous.
Thus, this dataset is not available to test whether an algorithm generally works well in the heterogeneous LiDAR sensor suites.
To tackle this problem, Qingqing~\etalcite{qingqing2022tiers} proposed the TIERS dataset, which consists of three omnidirectional LiDAR sensors and three solid-state LiDAR sensors.
Similar to the TIERS dataset, Jung~\etalcite{jung2023helipr} proposed the HeLiPR dataset, which is acquired by two omnidirectional LiDAR sensors and two solid-state LiDAR sensors, including under-researched channels, \ie~reflectivity, near-infrared, and radial velocity.
Unfortunately, these datasets only aim to evaluate odometry and place recognition, without any point-wise labels, as summarized in \tabref{table:summary}.

Regarding perception, Behley~\etalcite{behley2019iccv} proposed the SemanticKITTI dataset, a pioneering work that first provides point-wise semantic, instance, and MOS labels for 3D sequential point clouds.
Inspired by the SemanticKITTI, Pan~\etalcite{pan2020semanticposs} proposed the SemanticPOSS dataset, which shares exactly the same labeling protocol with SemanticKITTI to support compatibility with existing SemanticKITTI dataloaders.
While these datasets provide abundant point-wise labels, the SemanticKITTI and SemanticPOSS are only captured by a single omnidirectional LiDAR sensor.

\definecolor{table1cellcolor}{rgb}{1.0,1.0,0.6}

\begin{table}[t!]
    \captionsetup{font=footnotesize}
    \centering
    \caption{Comparison between existing 3D point cloud datasets and our proposed dataset. The term \textit{Hetero} indicates whether a dataset comprises both mechanically spinning omnidirectional and solid-state LiDAR sensors. We consider 2D and 3D omnidirectional LiDAR sensors to be homogeneous to each other.
    The symbol \trimark$\,$ indicates that the dataset provides point-wise labels; however, it incorrectly labels parked vehicles as moving objects by na\"{i}vely considering all pedestrians and vehicles as in motion.}
{\scriptsize
        \begin{tabular}{l|l|ccccc}
            \toprule \midrule
            & Dataset & Year & \begin{tabular}{@{}c@{}}Multiple \\ LiDARs\end{tabular} & Hetero & \begin{tabular}{@{}c@{}}Point-wise\\ MOS labels\end{tabular} \\ \midrule
            \parbox[t]{5mm}{\multirow{8}{*}{\rotatebox[origin=c]{90}{ \begin{tabular}{@{}c@{}}Odometry \& \\ place recognition \end{tabular}}}} &
            KITTI~\cite{geiger2012cvpr} & 2012 & \xmark  & \xmark & \xmark  \\
            & NCLT~\cite{carlevaris2016nclt} & 2016 & \xmark   & \xmark & \xmark  \\
            & Oxford Robotcar~\cite{maddern2017oxford} & 2017 & \cmark   & \xmark & \xmark  \\
            & Complex Urban~\cite{jeong2019complex} & 2019 & \cmark   & \xmark & \xmark  \\
            & MulRan~\cite{kim2020mulran} & 2020 & \xmark   & \xmark & \xmark \\
            & LIBRE~\cite{carballo2020libre} & 2020 & \cmark  & \xmark & \xmark \\
            & TIERS~\cite{qingqing2022tiers} &2022 & \cmark   & \cmark & \xmark \\
            & HeLiPR~\cite{jung2023helipr} & 2023 & \cmark & \cmark    & \xmark \\ \midrule
            \parbox[t]{0mm}{\multirow{8}{*}{\rotatebox[origin=c]{90}{Perception}}} &
            KITTI~\cite{geiger2012cvpr} & 2012 & \xmark  & \xmark & \xmark  \\
            & SemanticKITTI~\cite{behley2019iccv} & 2019 & \xmark & \xmark & \cmark  \\
            & SemanticPOSS~\cite{pan2020semanticposs} & 2020 & \xmark & \xmark & \trimark  \\
            & nuScenes~\cite{caesar2020cvpr} & 2020 & \xmark  & \xmark & \xmark   \\
            & WOMD~\cite{ettinger2021womd} & 2021 & \cmark & \cmark & \xmark   \\
            & PandaSet~\cite{xiao2021pandaset} & 2021 & \cmark  & \cmark & \xmark  \\
            & WOMD-LiDAR~\cite{chen2023womd} & 2023 & \cmark & \cmark & \xmark \\
            & \cellcolor{table1cellcolor}HeLiMOS (Ours) & \cellcolor{table1cellcolor}2024 & \cellcolor{table1cellcolor}\cmark & \cellcolor{table1cellcolor}\cmark & \cellcolor{table1cellcolor}\cmark \\
            \midrule \bottomrule
        \end{tabular}
    }
    \label{table:summary}
\end{table}
 
In recent years, Caesar~\etalcite{caesar2020cvpr} proposed the nuScenes dataset, which supports various perception tasks in 1,000 sequences.
Ettinger~\etalcite{ettinger2021womd}, Xiao~\etalcite{xiao2021pandaset}, and Chen~\etalcite{chen2023womd} proposed the WOMD, PandaSet, and WOMD-LiDAR datasets, respectively, which contain point clouds from multiple heterogeneous LiDAR sensors.
However, these datasets are also inappropriate to evaluate the performance of MOS in the heterogeneous LiDAR sensor setups because they do not provide point-wise MOS labels.
Therefore, to the best of our knowledge, we first propose a point-wise MOS dataset for heterogeneous LiDAR sensors, enabling the evaluation of MOS and static map building tasks across diverse LiDAR sensor setups.

Furthermore, we propose an efficient instance-aware automatic labeling framework to substantially lessen the annotation burden of a human labeler.
It is challenging and time-consuming for human labelers to discern moving objects in the 3D point clouds owing to the sparse characteristics of 3D point clouds~\cite{gebrehiwot2022teachers}.
To account for this, Kim and Kim~\cite{kim2020iros} proposed Removert, which is a range image-based scan-wise MOS labeling approach.
Furthermore, Chen~\etalcite{chen2022ral} proposed an automatic labeling framework called Auto-MOS.
In contrast to these prior approaches, we take instance information into account to reduce the number of false positives and thus minimize the need for manual corrections by a human labeler.
Thus, we propose instance-aware MOS annotation using ERASOR2, while accounting for the pose uncertainty in the revisited scenes via a topology-based trajectory clustering approach.

\section{Instance-Aware Automatic Labeling\\ and Data Statistics}
\label{sec:main}

\begin{figure}[t]
    \captionsetup{font=footnotesize}
    \centering
    \begin{subfigure}[b]{0.11\textwidth}
        \centering
        \includegraphics[width=1.0\textwidth]{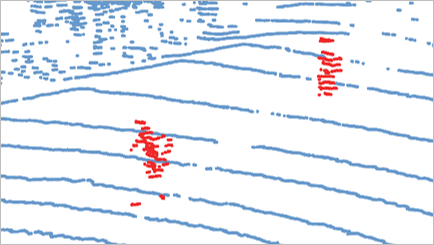}
        \includegraphics[width=1.0\textwidth]{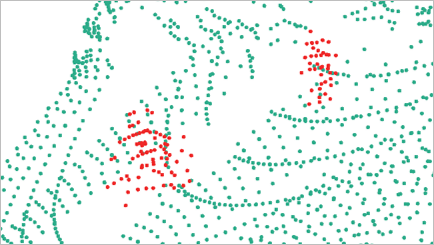}
        \includegraphics[width=1.0\textwidth]{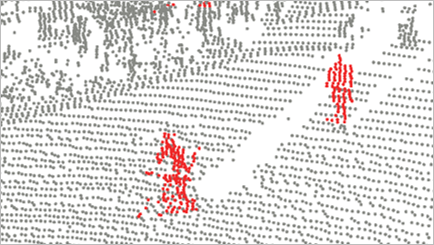}
        \includegraphics[width=1.0\textwidth]{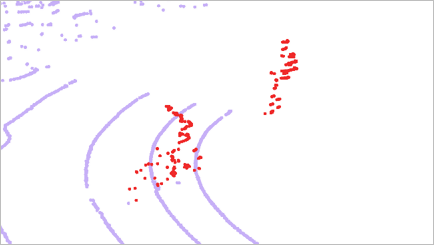}
        \caption{}
    \end{subfigure}
    \begin{subfigure}[b]{0.11\textwidth}
        \centering
        \includegraphics[width=1.0\textwidth]{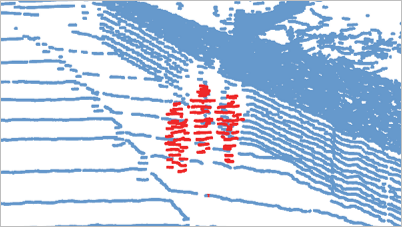}
        \includegraphics[width=1.0\textwidth]{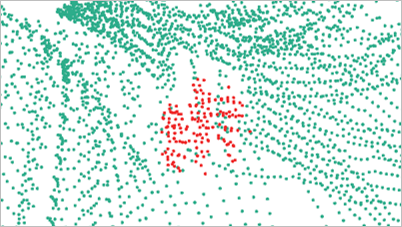}
        \includegraphics[width=1.0\textwidth]{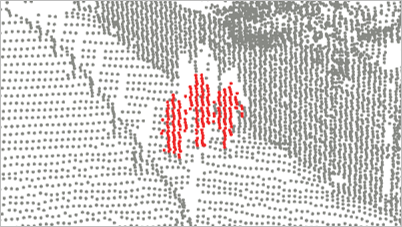}
        \includegraphics[width=1.0\textwidth]{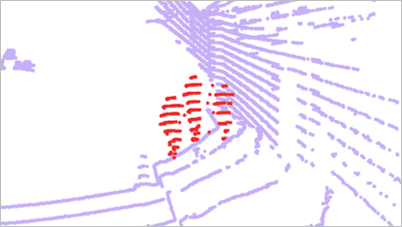}
        \caption{}
    \end{subfigure}
    \begin{subfigure}[b]{0.11\textwidth}
        \centering
        \includegraphics[width=1.0\textwidth]{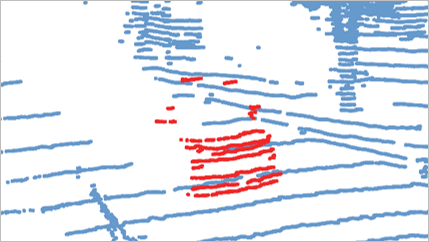}
        \includegraphics[width=1.0\textwidth]{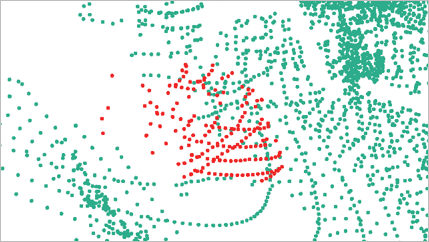}
        \includegraphics[width=1.0\textwidth]{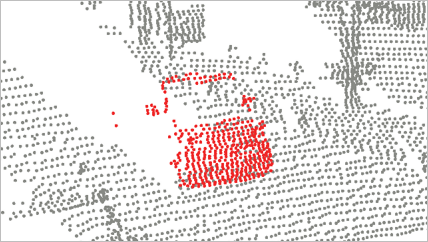}
        \includegraphics[width=1.0\textwidth]{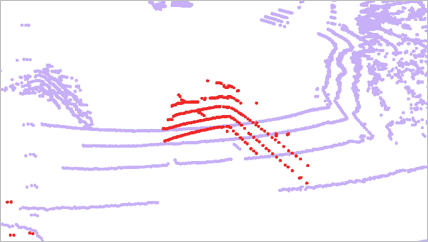}
        \caption{}
    \end{subfigure}
     \begin{subfigure}[b]{0.11\textwidth}
        \centering
        \includegraphics[width=1.0\textwidth]{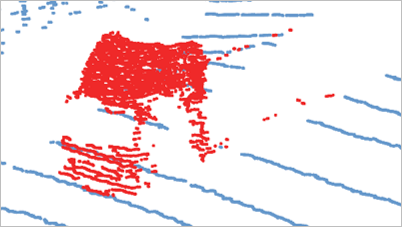}
        \includegraphics[width=1.0\textwidth]{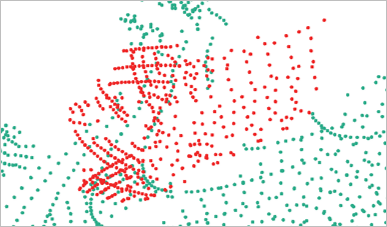}
        \includegraphics[width=1.0\textwidth]{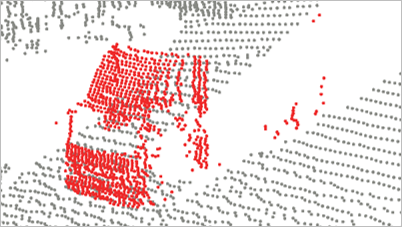}
        \includegraphics[width=1.0\textwidth]{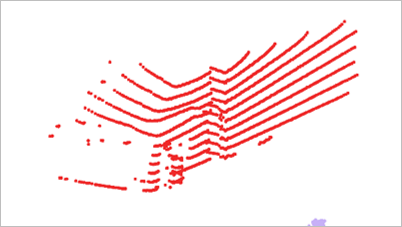}
        \caption{}
    \end{subfigure}
    \vsfigu
    \caption{Examples of moving objects in our dataset, which are shown as red points. From top to bottom, these examples show the zoomed point clouds captured by Aeva Aeries~\rom{2}, Livox Avia, Ouster OS2-128, and Velodyne VLP-16. Note that even though the same objects are shown, they have different patterns owing to the difference in scanning techniques and field of views of the sensors. MOS labels of (a)~a bicyclist and pedestrian, (b)~crowded pedestrians, (c)~a car, and (d)~a truck~(best viewed in color).}
    \label{fig:heli_lidar_viz}
    \vsfig
\end{figure}

Our dataset is based on the \texttt{KAIST05} sequence of HeLiPR dataset~\cite{jung2023helipr}, which contains various moving objects, such as buses, pedestrians, bicyclists, and cars, different from other sequences~(see \figref{fig:heli_lidar_viz}).
The dataset is acquired by four LiDAR sensors: Velodyne VLP-16 and Ouster OS2-128 as omnidirectional LiDAR sensors, and Livox Avia and Aeva Aeries~\rom{2} as solid-state LiDAR sensors, which have irregular scanning patterns compared with the traditional spinning LiDAR sensors.
For brevity, we denote these sensors as Velodyne~(\texttt{V}), Ouster~(\texttt{O}), Livox~(\texttt{L}), and Aeva~(\texttt{A}) in this paper, respectively.

Our goal is to provide a point-wise label for each point in the point clouds of all the LiDAR sensors.
Thus, we propose a merging-and-splitting-based efficient automatic MOS labeling framework, as illustrated in \figref{fig:overview}.
Our approach mainly consists of four steps.
First, we accumulate four point clouds from the four LiDAR sensors whose time steps are closest to each other by transforming them into the Ouster frame. By doing so, we synchronize the point clouds of these four LiDAR sensors at a software level, which is denoted by $\pi(\cdot)$ in \figref{fig:overview}.
In addition, we represent the accumulated point cloud by $\mathcal{P}_\curridx$ in \figref{fig:overview}(a).
Second, initial MOS labels are automatically annotated by our proposed automatic labeling framework, as presented in Figs.~\ref{fig:overview}(b) to (d).
Third, we manually correct the labels under human supervision.
Fourth, we backpropagate the refined MOS labels to the individual point clouds, as depicted in Figs.~\ref{fig:overview}(f) and (g).
The details are explained in the following subsections.

\subsection{Topology-Based Trajectory Clustering and Submap-Based Pose Correction}\label{section:tbtc}

In recent static map building approaches~\cite{lim2021ral, lim2023erasor2}, discrepancies in geometry or occupancy between individual scans and the map have often been used to estimate the dynamic points in the scans.
However, these approaches heavily rely on the assumption that the given poses are accurate and thus the scans are sufficiently well-aligned with each other.
Unfortunately, we have found that even though provided~(near) ground truth poses are used, undesirable errors exist in the poses for revisited scenes, \ie~loop-closed scenes.
These pose errors probably stem from systematic GNSS errors or potential errors arising from the process of aligning four point clouds because the point clouds were not originally synchronized at the hardware level.
Consequently, these errors make automatic labeling incorrectly classify static points as dynamic points, leading to many false positives and false negatives.

\newcommand{\cluster}{\mathcal{C}}
\begin{figure*}[t!]
	\centering
	\captionsetup{font=footnotesize}
	\begin{subfigure}[b]{0.90\textwidth}
		\centering
		\includegraphics[width=1.0\textwidth]{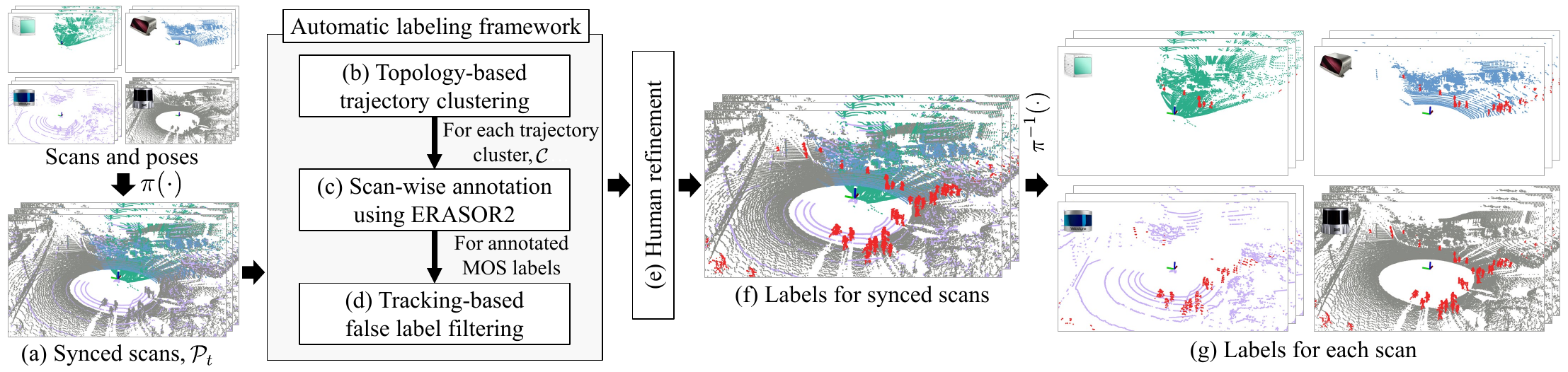}
	\end{subfigure}
	\vsfigu
	\caption{Overview of our merging-and-splitting-based labeling framework. (a)~Synchronization of the point clouds from the four LiDAR sensors at a software level. (b)-(d)~Procedure of our proposed automatic labeling framework.
		(b) First, trajectories are segmented into multiple clusters.
		(c)~For each trajectory cluster $\cluster$, we apply an instance-aware static map building, ERASOR2~\cite{lim2023erasor2}, that produces initial scan-wise annotated labels.
		(d)~Tracking-based false label filtering is applied to reduce false positive and false negative MOS labels.
		(e)~Next, these labels are manually corrected under human supervision.
		(f)-(g)~Finally, the refined labels of synced scans are backpropagated to individual point clouds, which is denoted by $\pi^{-1}(\cdot)$.
	    Red points indicate the annotated dynamic points~(best viewed in color).}
	\label{fig:overview}
	\vsfig
\end{figure*}

\begin{figure}[t]
	\centering
	\begin{subfigure}[b]{0.48\textwidth}
		\centering
		\includegraphics[width=1.0\textwidth]{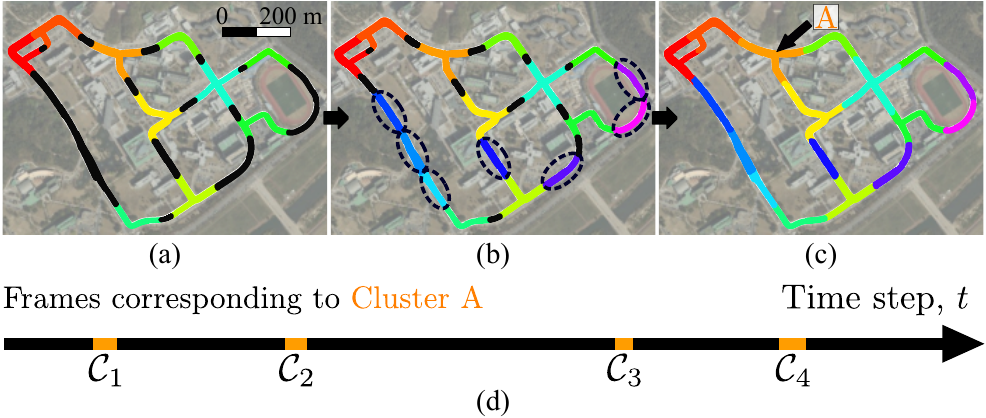}
	\end{subfigure}
	\captionsetup{font=footnotesize}
	\vsfig
	\caption{(a)-(c) Procedure of our topology-based trajectory clustering. Black trajectory indicates unclustered frames and each color represents a different cluster~(best viewed in color).
		(a)~First, intersections are prioritized because these scenes are highly likely to have multiple revisits.
		(b)~Next, \changed{the frames from} revisited places that are not intersections and \changed{consecutive frames without revisits but with sufficiently large intervals} are clustered \changed{respectively}, as indicated by the black dashed circles.
		(c)~Each unclustered frame is merged into the adjacent cluster with the closest frame interval.
		(d)~Frames included in Cluster A, which is indicated in (c), visualized along the time step axis. As a result of the clustering, several sets of consecutive frames are clustered together.}
	\label{fig:clustering}
	\vsfig
\end{figure}

To address this issue, we divide the trajectory with poses corresponding to $\mathcal{P}_t$ into multiple clusters and correct their poses to align their reference frames.
The positions of the trajectory are neither dense nor have geometrical features, making existing clustering methods not work~\cite{mcinnes2017hdbscan}.
For this reason, as presented in \figref{fig:clustering}, we propose topology-based trajectory clustering that prioritizes revisited sections, which are likely to have inherent pose errors owing to the significant time differences between scans captured during initial visits and those upon revisiting.
This is because time discrepancies can lead to pose drift, which may not be fully minimized even after pose graph optimization.

As illustrated in \figref{fig:clustering}, our trajectory clustering follows three steps.
First, we identify areas, such as intersections or places where left/right turns occur, by examining the yaw differences within the trajectory and then group neighboring frames into a cluster based on their position values as inputs.
Second, \changed{the frames from} revisited places that are not intersections and \changed{consecutive frames without revisits but with sufficiently large intervals} are clustered, \changed{respectively}.
Finally, the remaining unclustered frames are merged into the adjacent cluster with the closest frame interval.

Next, poses corresponding to frames within the same cluster are corrected to minimize errors between the reference frames for each subcluster.
Formally, let $\mathcal{C}$ be a cluster of the trajectory and the $n$-th consecutive frame set (or a subcluster) in $\mathcal{C}$ be~$\mathcal{C}_n$, which satisfies $\mathcal{C} = \bigcup_{n=1}^{N_c} \mathcal{C}_n$, as visualized in \figref{fig:clustering}(d); $N_c \geq 1$ denotes the number of the subclusters.
Based on the assumption that the poses in $\mathcal{C}_n$ are locally consistent,
we define the reference frames of each subcluster as different by denoting the transformation matrix of the $\curridx$-th body frame with respect to the reference frame~$w_n$ of the $n$-th subcluster as $\mathbf{T}^{w_n}_\curridx$.
Then, the $n$-th submap $\mathcal{S}_n$, which corresponds to $\mathcal{C}_n$, is defined as follows:

\vsequ
\begin{equation}
	\mathcal{S}_n=\nu \bigg(\bigcup_{\curridx\in \mathcal{C}_n}{\nu\Big(\big\{\mathbf{T}^{w_n}_\curridx \mathbf{p} \mid \mathbf{p} \in \mathcal{P}_\curridx\big\}}\Big) \bigg),
	\label{eq:accum_map}
\end{equation}
\vseq

\noindent where $\nu(\cdot)$ denotes a voxel sampling function with the voxel size $\nu_\text{map}$, $\mathcal{P}_{\curridx}$ is the synced scan whose origin is the $\curridx$-th body frame, and $\mathbf{T}^{w_n}_t \mathbf{p}$ means that a point $\mathbf{p}$ is transformed into the reference frame~$w_n$.

Finally, to locally unify the coordinate system into the reference frame of~$\mathcal{S}_1$, \ie~$w_1$ frame, we perform submap-to-submap ICP between $\mathcal{S}_1$ and $\mathcal{S}_n$ to estimate relative transformation $\hat{\mathbf{T}}^{w_1}_{w_n}$ and all the poses of $\mathcal{C}_n$ are updated as $\hat{\mathbf{T}}^{w_1}_{t} = \hat{\mathbf{T}}^{w_1}_{w_n} \mathbf{T}^{w_n}_{t}$, respectively.
Thus, ICP is performed $N_c - 1$ times for each cluster.

\subsection{Instance-Aware Initial Data Annotation}

Next, by taking the corrected poses and corresponding synced scans of $\mathcal{C}$ as inputs,
our instance-aware annotation pipeline is applied to generate initial scan-wise MOS labels by utilizing instance segmentation information~\cite{lim2023erasor2}, which corresponds to \figref{fig:overview}(c).
The previous automatic labeling approach by Chen~\etalcite{chen2022ral} employed ERASOR~\cite{lim2021ral} to initially annotate MOS labels and then clustering was applied, which was referred to as a \textit{detect-then-cluster} scheme.
As ERASOR does not account for instance information, it potentially fails to reject whole dynamic points from a moving object, considering some partial dynamic points as static.

In contrast, \changed{ERASOR2~\cite{lim2023erasor2} is a }\textit{cluster-then-detect} approach\changed{. That is, ERASOR2} first performs instance segmentation\changed{, followed by dynamic point removal at the instance level by checking geometrical discrepancies between each scan and the map cloud to determine which regions are temporarily occupied.}
By doing so, we can generate more accurate and reliable MOS labels.

\begin{figure}[t]
	\centering
	\begin{subfigure}[b]{0.46\textwidth}
		\centering
		\includegraphics[width=1.0\textwidth]{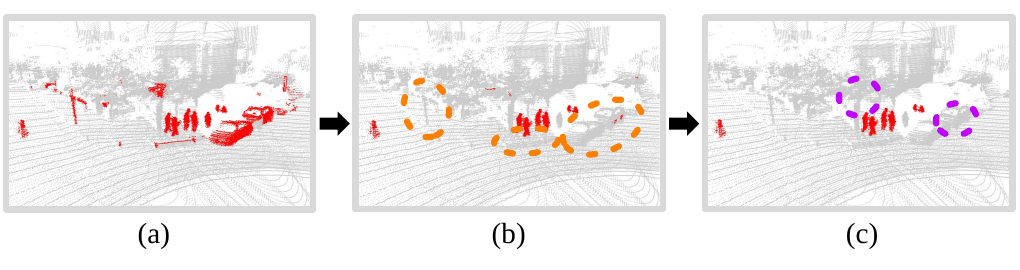}
	\end{subfigure}
	\captionsetup{font=footnotesize}
	\vsfigu
	\caption{(a)-(c) The annotation results in our proposed labeling framework.
	Red points denote the annotated dynamic points, while gray points represent points estimated to be static~(best viewed in color).
		(a)~The initial result obtained by using ERASOR2. (b)~Refined annotation through our tracking-based filtering. Orange dashed circles indicate that false positive points are successfully rejected.
		(c)~Final annotation after human supervision. Purple dashed circles highlight the refined areas by a human labeler.}
	\label{fig:labeling_procedure}
	\vsfig
\end{figure}

\begin{figure*}[t!]
    \centering
    \begin{subfigure}[b]{0.22\textwidth}
        \centering
        \includegraphics[width=1.0\textwidth]{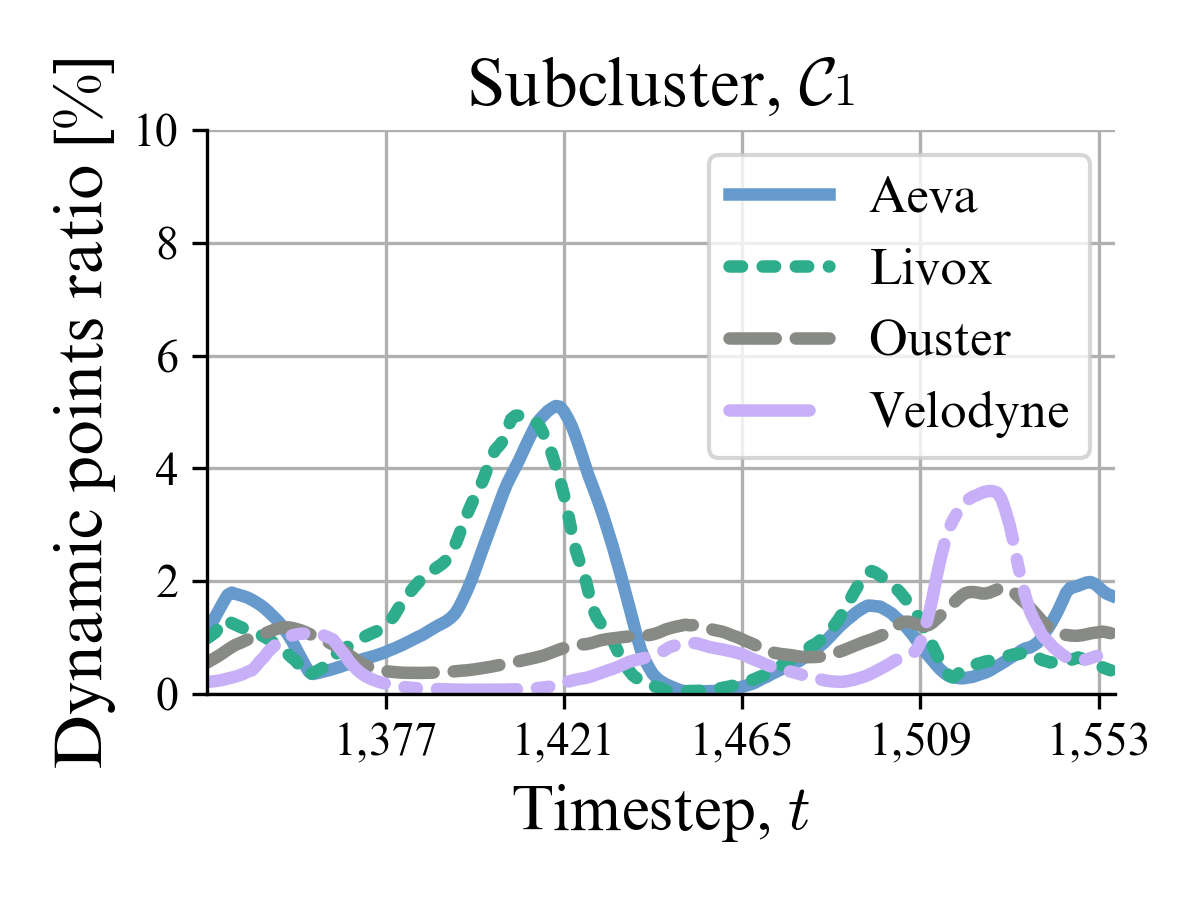}
        \includegraphics[width=1.0\textwidth]{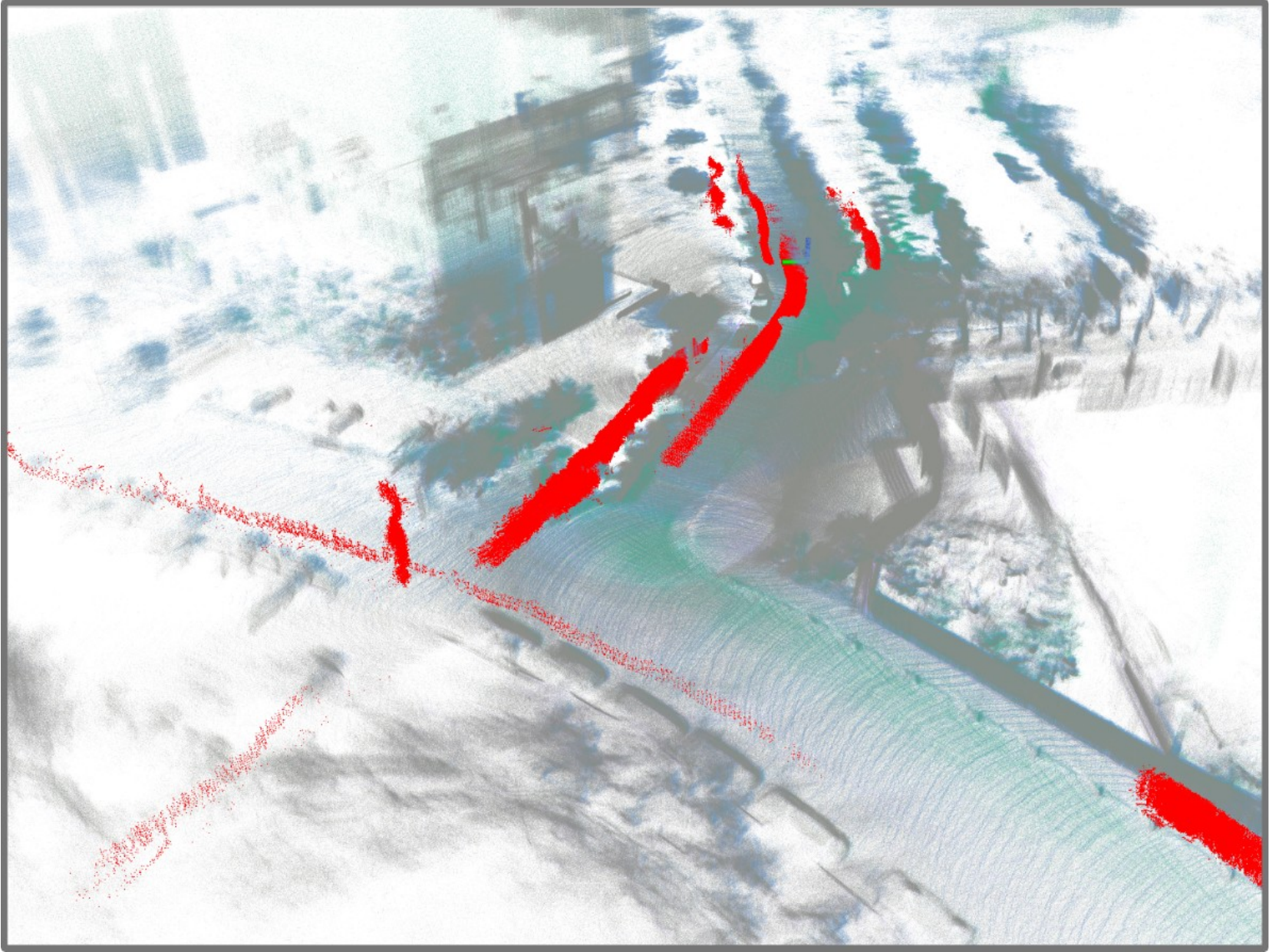}
    \end{subfigure}
    \begin{subfigure}[b]{0.22\textwidth}
        \centering
        \includegraphics[width=1.0\textwidth]{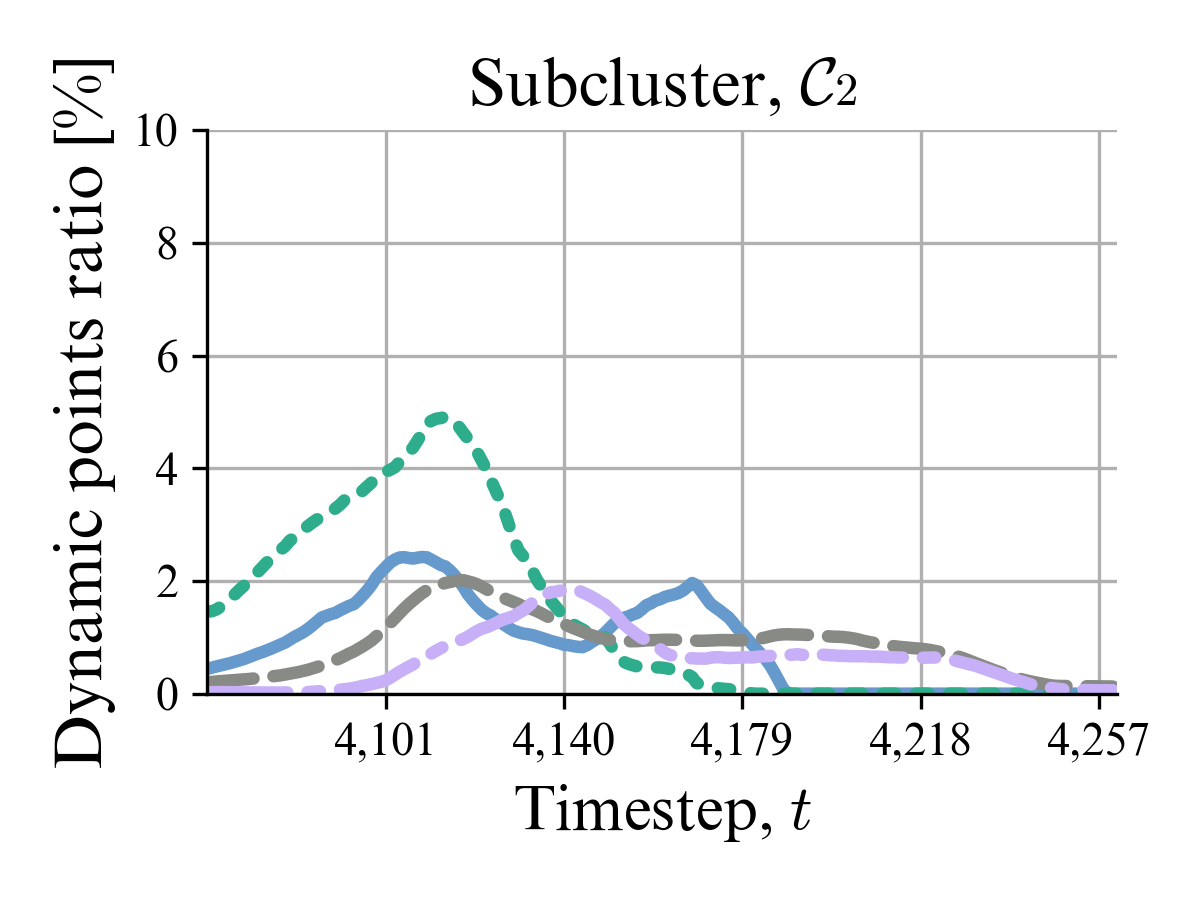}
        \includegraphics[width=1.0\textwidth]{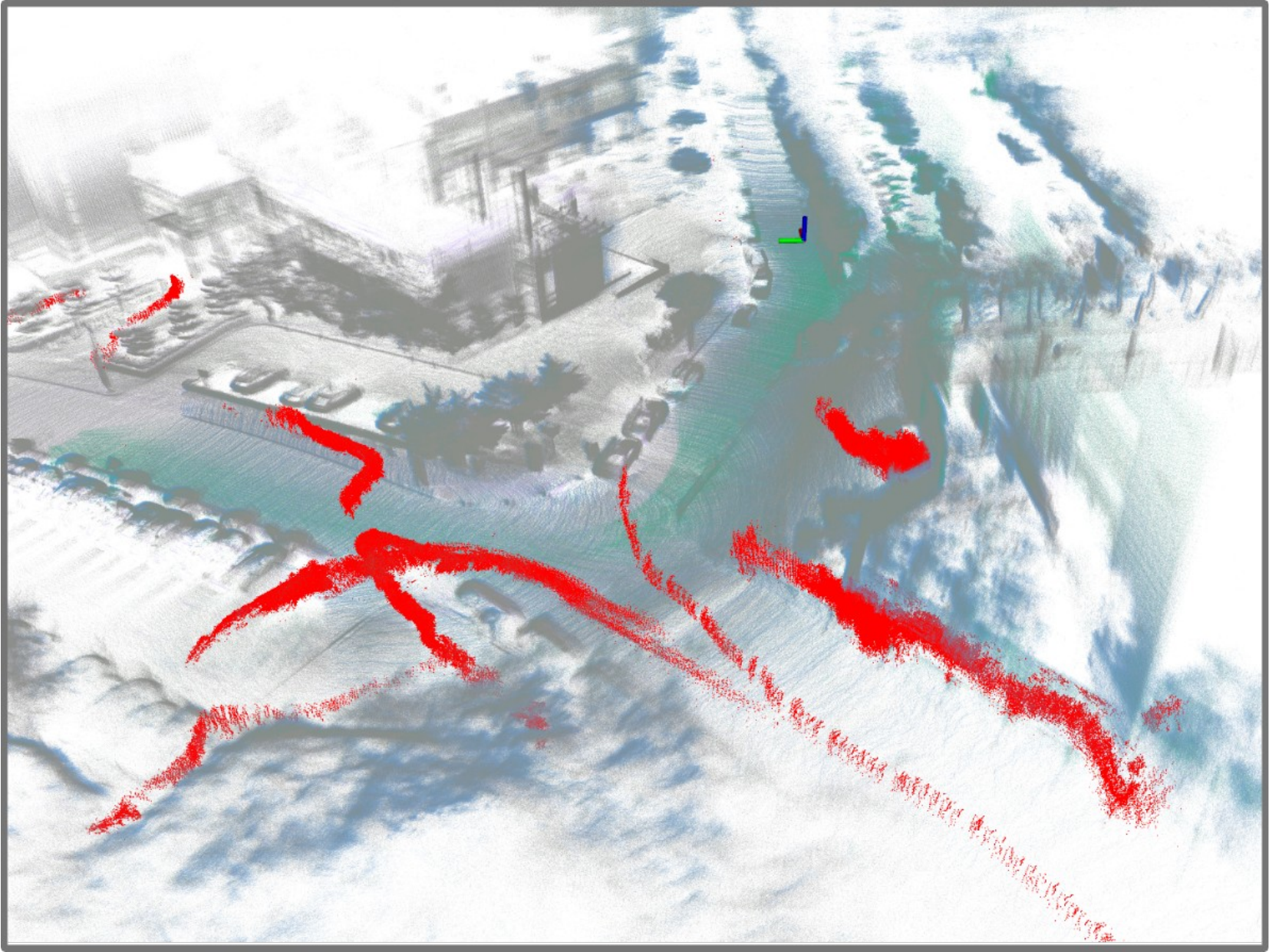}
    \end{subfigure}
    \begin{subfigure}[b]{0.22\textwidth}
        \centering
        \includegraphics[width=1.0\textwidth]{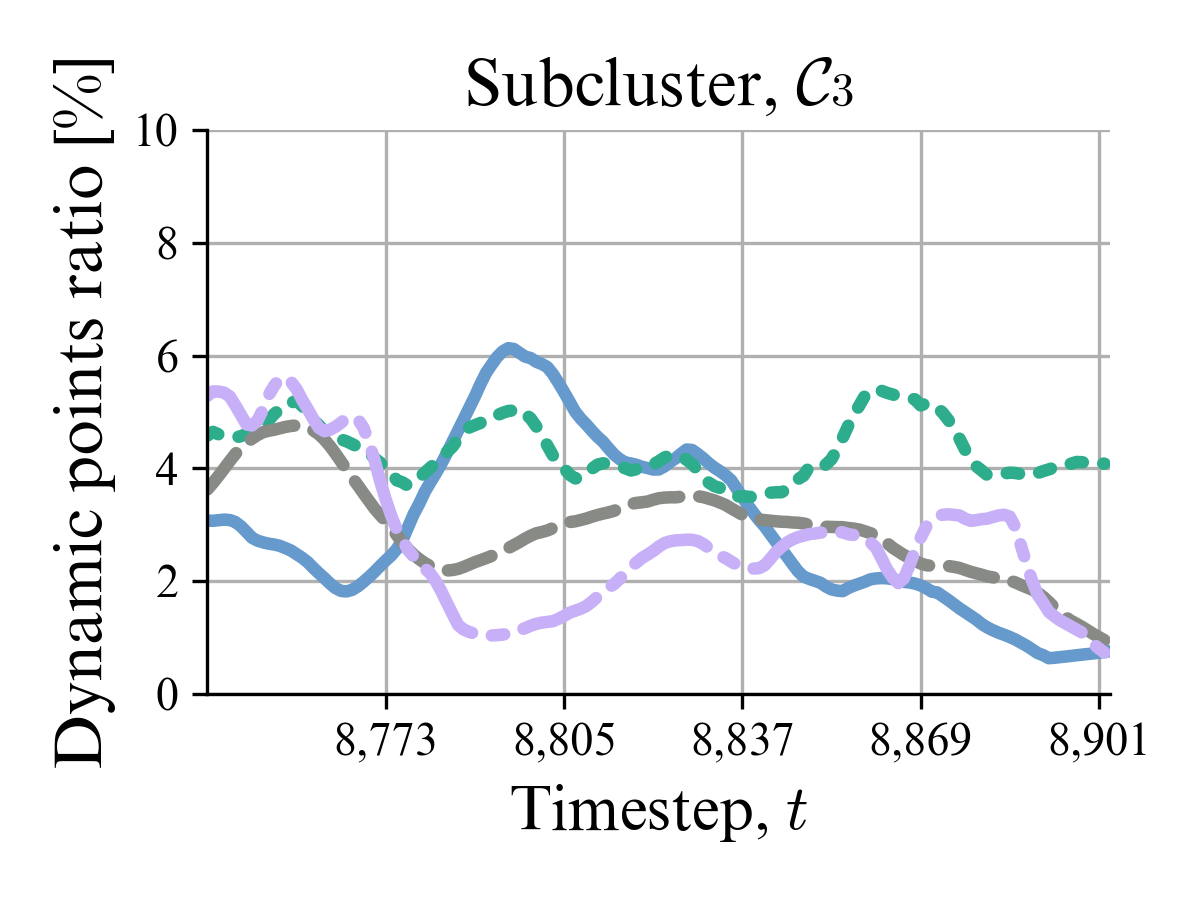}
        \includegraphics[width=1.0\textwidth]{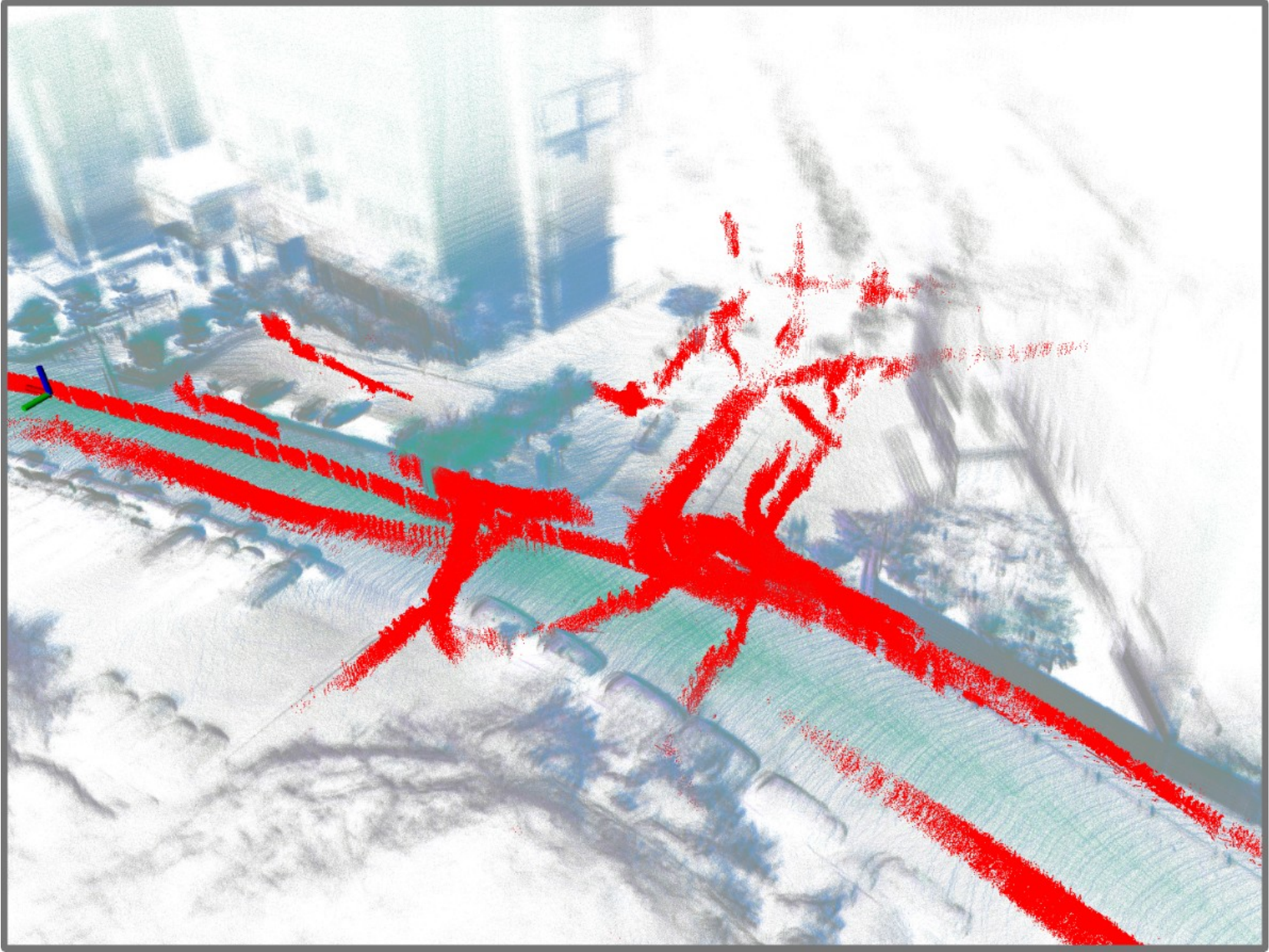}
    \end{subfigure}
    \begin{subfigure}[b]{0.22\textwidth}
        \centering
        \includegraphics[width=1.0\textwidth]{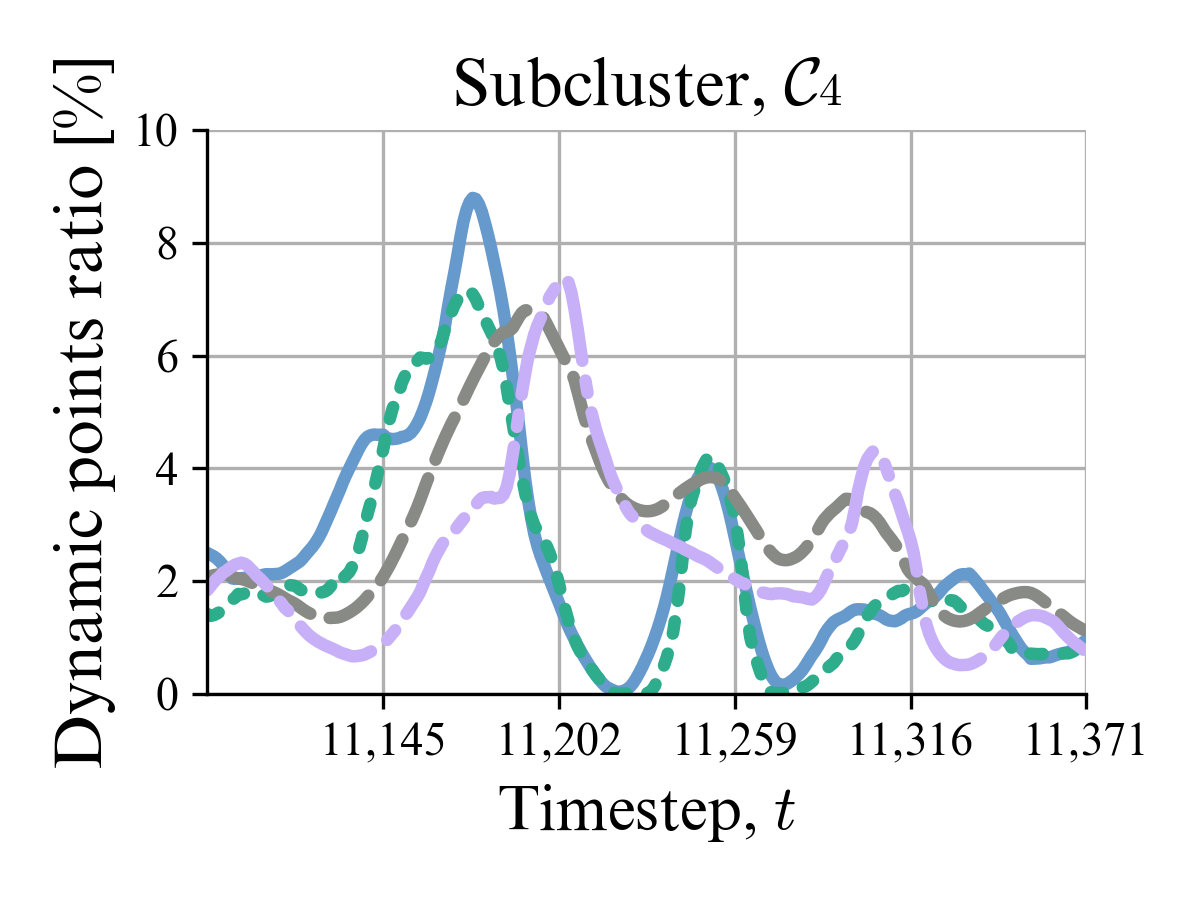}
        \includegraphics[width=1.0\textwidth]{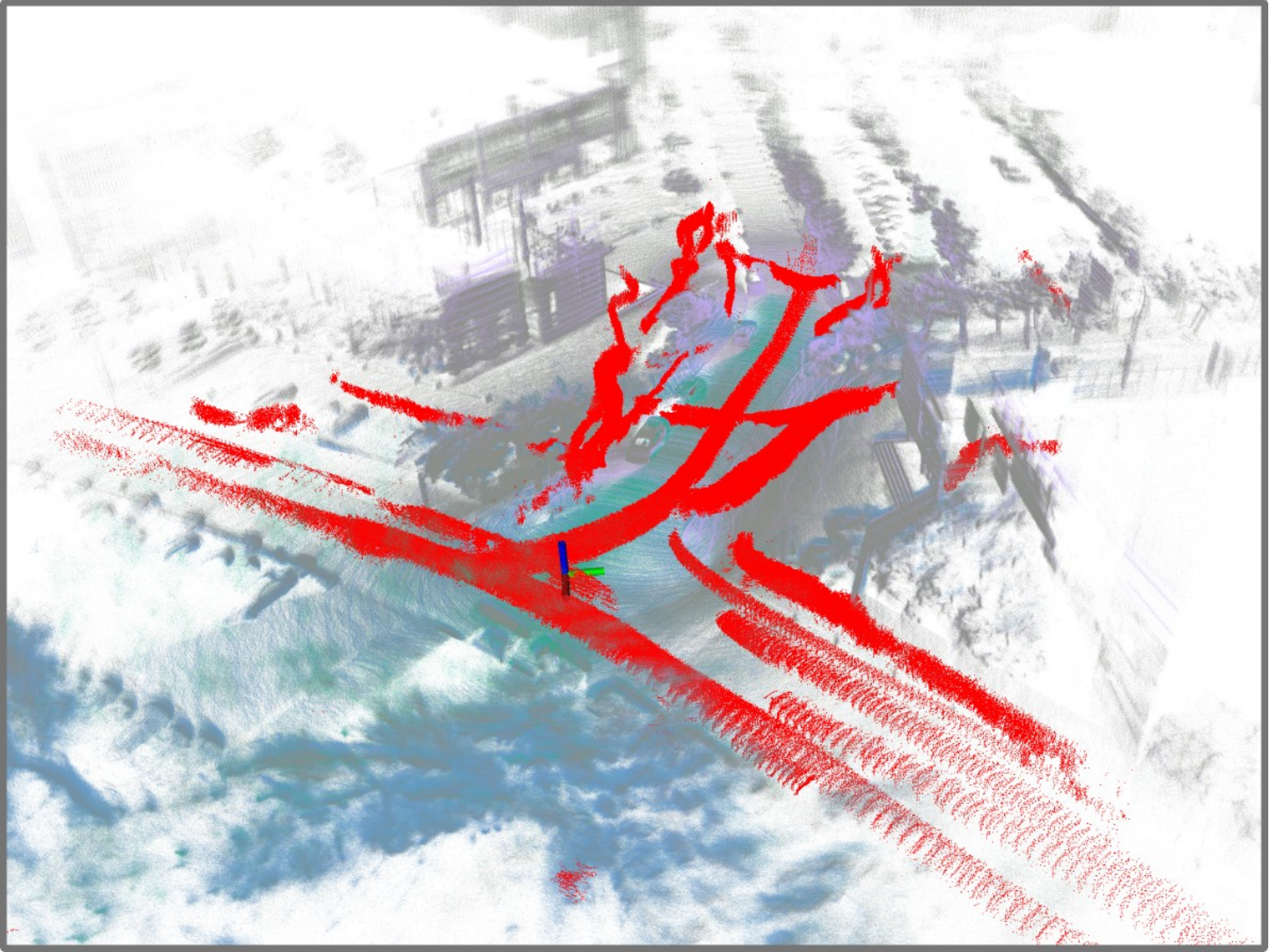}
    \end{subfigure}
    \captionsetup{font=footnotesize}
    \caption{(T-B, L-R):~Dynamic points ratios, each of which is defined as $\frac{\text{\# of labeled dynamic points}}{\text{\# of total points in the $t$-th scan}}$, over time steps and the visualized cleaned point cloud maps of four subclusters, corresponding to $\mathcal{C}_1$, $\mathcal{C}_2$, $\mathcal{C}_3$, and $\mathcal{C}_4$ in \figref{fig:clustering}(d).
    \changed{Note that} the MOS labels of our HeLiMOS show a variety of dynamic point patterns owing to the varying trajectories of moving objects even though the scans are acquired in the same places.
    \changed{This is because the point clouds in our dataset are from the HeLiPR dataset which is originally designed for place recognition tasks to evaluate whether a robot revisits the same location; thus, our dataset also includes scans captured at the same locations multiple times at different moments.}
    Red points indicate the annotated points, which are traces of moving objects~(best viewed in color).}
    \label{fig:data_statistics}
    \vsfig
\end{figure*}
 
\begin{figure}[t]
  \centering
  \begin{subfigure}[b]{0.44\textwidth}
    \centering
    \includegraphics[width=1.0\textwidth]{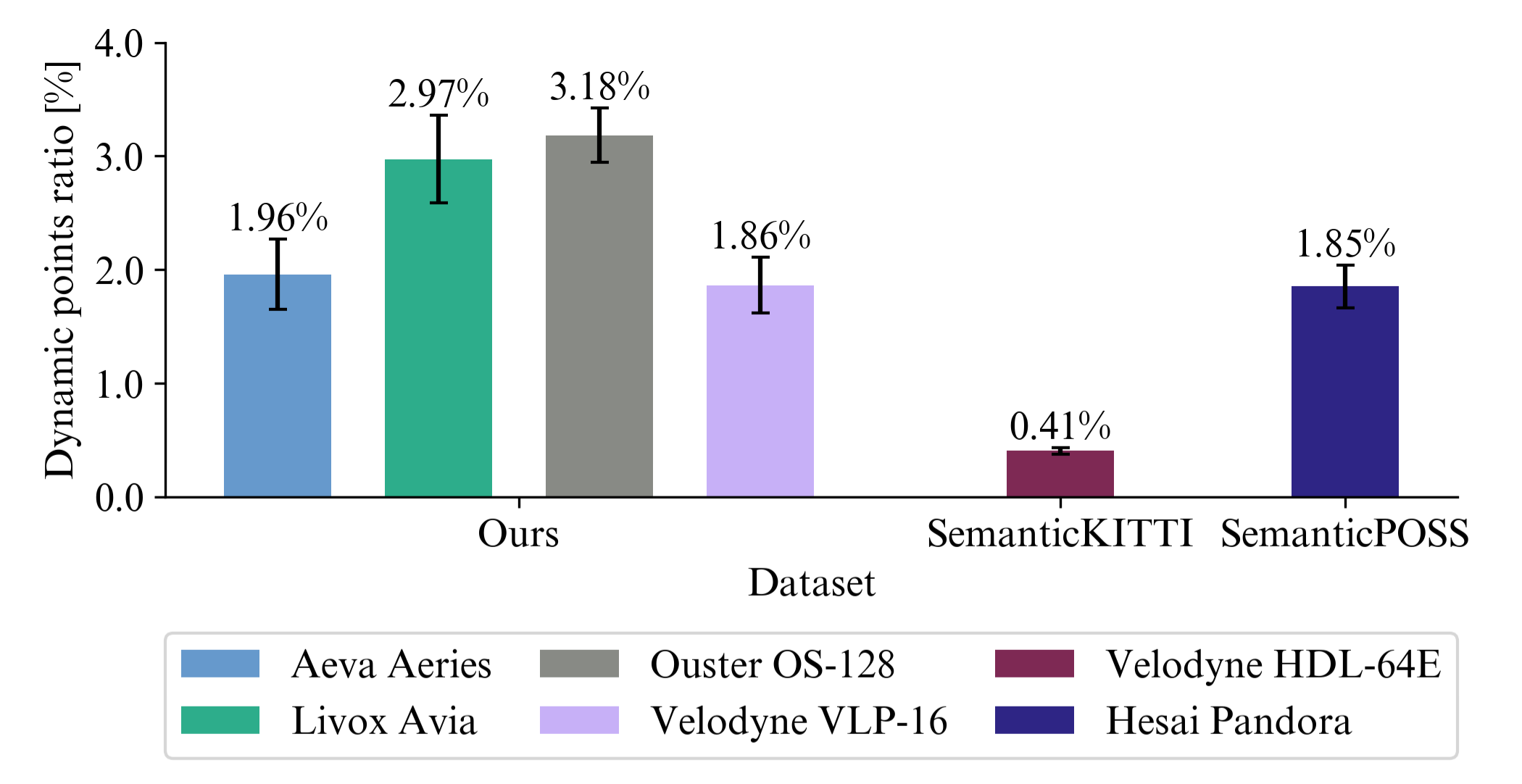}
    \end{subfigure}
    \captionsetup{font=footnotesize}
    \caption{Comparison of dynamic points ratio with other datasets. The numbers on the bars represent the average ratios, while the black error bars indicate the standard deviations.
    For calculating the dynamic points ratio of SemanticKITTI~\cite{behley2019iccv}, we counted the points labeled as moving objects.
    As described in \tabref{table:summary}, the SemanticPOSS~\cite{pan2020semanticposs} wrongly classifies parked vehicles as moving objects.
    Thus, for a fair comparison, we filtered out them using our tracking-based filtering and only used the actual moving objects for the dynamic points ratio calculation.}
  \label{fig:heli_kitti_poss}
  \vsfig
\end{figure}

\subsection{Multi-Object Tracking-Based False Label Filtering and Human Refinement}\label{sec:tracking_based_filtering}

The aforementioned static map building approach-based automatic labeling is likely to remove dynamic points somewhat aggressively because static map building approaches were originally designed to preserve definite static points for performing localization or navigation.
For this reason, as presented in \figref{fig:labeling_procedure}(a), many static points are wrongly classified as dynamic points at the scan level.
To address this issue, we leverage multi-object tracking-based filtering~\cite{jang2023toss}.
In contrast to Chen~\etalcite{chen2022ral}, who also employed tracking-based filtering primarily focusing on the rejection of false positive points,
we propose a bounding box augmentation to reduce the number of false negative points.
That is, we augment additional bounding boxes in the frames where tracking is temporarily lost by interpolating the centroids of bounding boxes tracked in the previous frame and next frame.
Subsequently, points within these augmented bounding boxes are also classified as dynamic points and thus are successfully rejected.
As a result, more refined MOS labels can be obtained without human effort, as shown in \figref{fig:labeling_procedure}(b).

Nevertheless, these procedures do not perfectly reject all false positives and negatives.
Therefore, as a final stage, we perform a human-in-the-loop refinement process to enhance the quality of the MOS labels, as depicted in \figref{fig:labeling_procedure}(c).

\subsection{Data Statistics and File Structure}\label{sec:data_statistics}

Our dataset provides a total of 12,188 labeled point clouds. Each MOS label follows the SemanticKITTI-MOS format, so it consists of three classes: \textit{unlabeled}, \textit{static}, and \textit{dynamic}.
\changed{The point clouds in our dataset are from the HeLiPR dataset, which is originally designed for place recognition tasks to evaluate whether a robot revisits the same location.}
\changed{For this reason,} our HeLiMOS inherits \changed{characteristics of the HeLiPR dataset} and \changed{thus} leverages two notable \changed{attributes}:
(a)~the inclusion of several revisited scenes and (b)~a significantly higher ratio of dynamic points.

\begin{figure}[t]
	\centering
	\begin{subfigure}[b]{0.45\textwidth}
		\centering
		\includegraphics[width=1.0\textwidth]{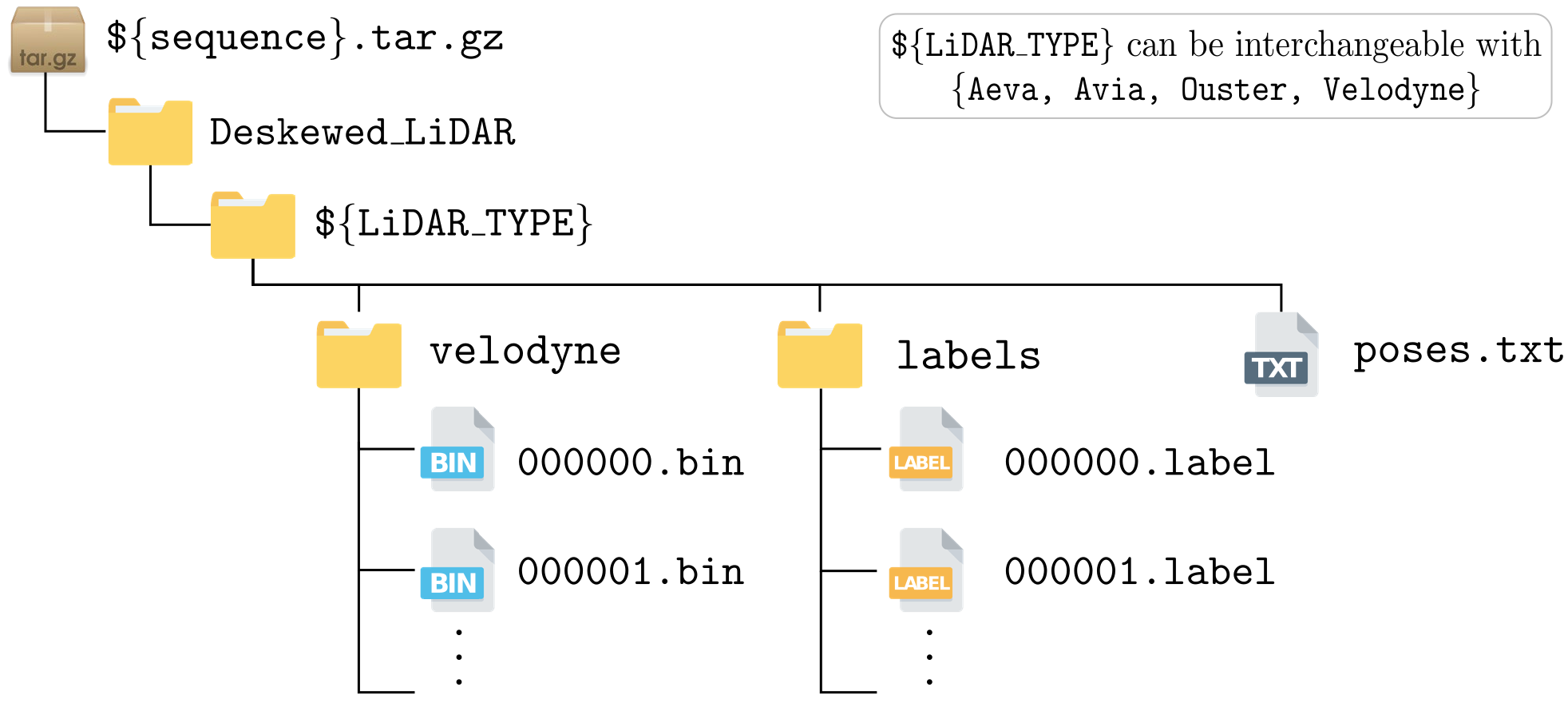}
	\end{subfigure}
	\captionsetup{font=footnotesize}
	\vsfigu
	\caption{File structure of our dataset, which follows the SemanticKITTI format~\cite{behley2019iccv}. It may seem awkward to use the folder name \textit{velodyne} instead of \textit{scans}, but we adhere to the convention of using \textit{velodyne} as it is used in other datasets, such as SemanticPOSS~\cite{pan2020semanticposs}. Pose information is from the original dataset~\cite{jung2023helipr}.}
	\label{fig:file_system}
	\vsfig
\end{figure}

As shown in Fig.~\ref{fig:data_statistics}, we can observe that dynamic points ratios and dynamic point patterns vary significantly over time, even though scans are acquired in the same place.
For instance, in the previously mentioned clusters, $\mathcal{C}_3$ and $\mathcal{C}_4$ contain more moving objects and more complex trajectory patterns, resulting in higher dynamic points ratios compared with $\mathcal{C}_{1}$ and $\mathcal{C}_{2}$.
In addition, owing to the different field of views of each sensor, the scanning patterns of static scenes also become different, as presented in \figref{fig:data_statistics}.
Because it is crucial for MOS approaches to robustly distinguish moving objects regardless of changes in the dynamic points ratios or moving object trajectory patterns,
our dataset can provide an opportunity to evaluate the generalization capabilities of MOS across diverse patterns in the same scene.

Furthermore, note that the most distinctive feature of our dataset is that it not only has higher dynamic points ratios than existing MOS datasets, but also has MOS labels of four heterogeneous LiDAR sensors.
As presented in Fig.~\ref{fig:heli_kitti_poss}, our dataset shows consistently higher average dynamic points ratios across all LiDAR sensors compared with the SemanticKITTI~\cite{behley2019iccv} and SemanticPOSS~\cite{pan2020semanticposs} datasets.

Therefore, by using our dataset, researchers can evaluate the generalization capabilities of MOS approaches against untrained environments and different types of LiDAR sensors.
As presented in \figref{fig:file_system}, the file structure of our dataset follows the SemanticKITTI format~\cite{behley2019iccv} to support compatibility with existing SemanticKITTI dataloaders.
All the laser scans were deskewed and then saved by utilizing HeLiPR Pointcloud Toolbox\footnote{\scriptsize\texttt{https://github.com/minwoo0611/HeLiPR-Pointcloud-Toolbox}}{}.
Next, we split the dataset into training, validation, and test sets with ratios of 68\%, 16\%, and 16\%, respectively.
Note that we do not randomly sample the frames; instead, we designate certain sequential frames from the revisited scenes, \eg~$\mathcal{C}_3$ or $\mathcal{C}_4$ in \figref{fig:data_statistics}, for the validation and test sets.

\section{Evaluation of Moving Object Segmentation and Static Map Building}
\label{sec:exp}

The main focus of this work is to provide point-wise MOS labels for evaluating the generalization capabilities of MOS in heterogeneous LiDAR sensor setups.
In addition, our dataset can be utilized to evaluate the performance of static map building approaches.
Thus, we present three experiments by utilizing our dataset:
(i)~MOS performance of the models trained on the SemanticKITTI dataset~\cite{behley2019iccv} against both environmental changes and LiDAR sensor type variations,
(ii)~MOS performance across heterogeneous LiDAR sensors, and (iii)~automatic labeling performance to support the rationale behind our choice to use ERASOR2 and the proposed tracking-based filtering.
These novel experiments, which could not be evaluated using existing datasets, back up our key claim of the necessity of the heterogeneous LiDAR MOS dataset and our automatic labeling framework.

\subsection{Experimental Setup}

\newcommand{\iou}{\text{IoU}_{\text{MOS}}}
\newcommand{\PRmap}{\text{PR}}
\newcommand{\RRmap}{\text{RR}}

In the first experiment, we use the pre-trained MOS models on the SemanticKITTI dataset~\cite{behley2019iccv}, which is captured by a 64-channel omnidirectional LiDAR sensor, and then quantitatively evaluate the inference results of the models by using all the labels.
In the second experiment, we train MOS approaches on one type of LiDAR sensors and then test on heterogeneous LiDAR sensors, \ie~training with point clouds from solid-state LiDAR and testing with those from omnidirectional LiDAR sensors, or vice versa, to examine performance variations across different LiDAR types.

As a quantitative metric, we use the intersection-over-union~(IoU) for MOS~\cite{mersch2022ral}, $\iou$, which is defined as follows:

\begin{equation}
    \iou=\frac{\mathrm{TP}}{\mathrm{TP}+\mathrm{FP}+\mathrm{FN}},
	\label{eqn:iou}
\end{equation}

\noindent where $\mathrm{TP}$, $\mathrm{FP}$, and $\mathrm{FN}$ denote the true positive, false positive, and false negative points from the perspective of MOS, respectively.

For the third experiment, we evaluate the modules of our labeling framework and existing approaches by using preservation rate~(PR), rejection rate~(RR), and F$_{1}$ score~\cite{lim2021ral,lim2023erasor2}, defined as:

\begin{itemize}
	\item $\PRmap = \textcolor{black}{\frac{\text{\# of preserved static voxels}}{\text{\# of total static voxels on the naively accumulated map}}}$,
	\item $\RRmap = \textcolor{black}{1- \frac{\text{\# of remaining dynamic voxels}}{\text{\# of total dynamic voxels on the naively accumulated map}}}$,
	\item $\text{F}_1 = 2\PRmap \cdot \RRmap / (\PRmap + \RRmap)$.
\end{itemize}

\noindent We assess the performance of the static map building approaches with synced scans, \ie~$\mathcal{P}_t$, as inputs.

For simplicity, we refer to each sensor type used in our dataset as \texttt{L}, \texttt{A}, \texttt{O}, and \texttt{V}, respectively, as described in \secref{sec:main}.

\subsection{Moving Object Segmentation Performance Against Environmental Changes and LiDAR Sensor Variations}

First, we evaluate the generalization capabilities of MOS approaches in untrained environments and the different types of LiDAR sensors.
To this end, we mainly employ 4DMOS~\cite{mersch2022ral} and MapMOS~\cite{mersch2023building}, which are state-of-the-art volumetric MOS approaches that do not employ range image projection and thus can be directly applied in other LiDAR setups.

We can analyze the results of this experiment in three aspects.
First, we demonstrate the robustness of these volumetric MOS approaches against environmental changes.
This is evidenced by the relatively little performance degradation with \texttt{O}, which is the most similar sensor to the 64-channel sensor used to acquire SemanticKITTI.
Second, in contrast, we observed substantial performance degradation in solid-state LiDAR cases.
Third, when using sparser point clouds as inputs,
the performance of MOS approaches was degraded more significantly~(see columns~\texttt{L} and \texttt{V} in \tabref{table:kitti2helimos}).
This is because these MOS approaches heavily depend on the pose estimation modules to use temporal information from LiDAR sequences.
Consequently, once the estimated poses are imprecise owing to the sparse point clouds, the MOS performance becomes worse.

\begin{table}[t!]
	\centering
	\captionsetup{font=footnotesize}
	\caption{Mean IoU of MOS approaches trained on the SemanticKITTI dataset to evaluate generalization capabilities in terms of both environmental changes and LiDAR sensor variations ($\texttt{L}$:~Livox Avia, $\texttt{A}$:~Aeva Aeries~\rom{2}, $\texttt{O}$:~Ouster OS2-128, and $\texttt{V}$:~Velodyne VLP-16).}
	{\scriptsize
	\begin{tabular}{lccccc}
		\toprule \midrule
		\multirow{2}{*}[-0.2em]{Method} & \multicolumn{2}{c}{Solid-state} & \multicolumn{2}{c}{Omnidirectional} & \multirow{2}{*}[-0.2em]{Total}  \\ \cmidrule(lr){2-3} \cmidrule(lr){4-5}
		  & \texttt{L} &  \texttt{A} & \texttt{O} & \texttt{V} \\ \midrule
4DMOS, online~\cite{mersch2022ral} & 52.08 & 54.01 & 64.17 & 4.69 & 43.74  \\
	  	 4DMOS, delayed~\cite{mersch2022ral} & 58.99 & 58.30 & 70.44 & 5.41 & 48.28 \\
		 MapMOS, Scan~\cite{mersch2023building} & 58.93 & 63.15 & 81.43 & 4.33 & 51.96 \\
	 MapMOS, Volume~\cite{mersch2023building} & \textbf{62.70} & \textbf{66.58} & \textbf{82.87} & \textbf{5.77} & \textbf{54.48} \\
		\midrule \bottomrule
	\end{tabular}
	}
	\label{table:kitti2helimos}
\end{table}

\subsection{Moving Object Segmentation Performance Across Heterogeneous LiDAR Sensors}

\begin{figure}[t!]
	\centering
	\captionsetup{font=footnotesize}
	\begin{subfigure}[b]{0.48\textwidth}
		\centering
		\includegraphics[width=0.23\textwidth]{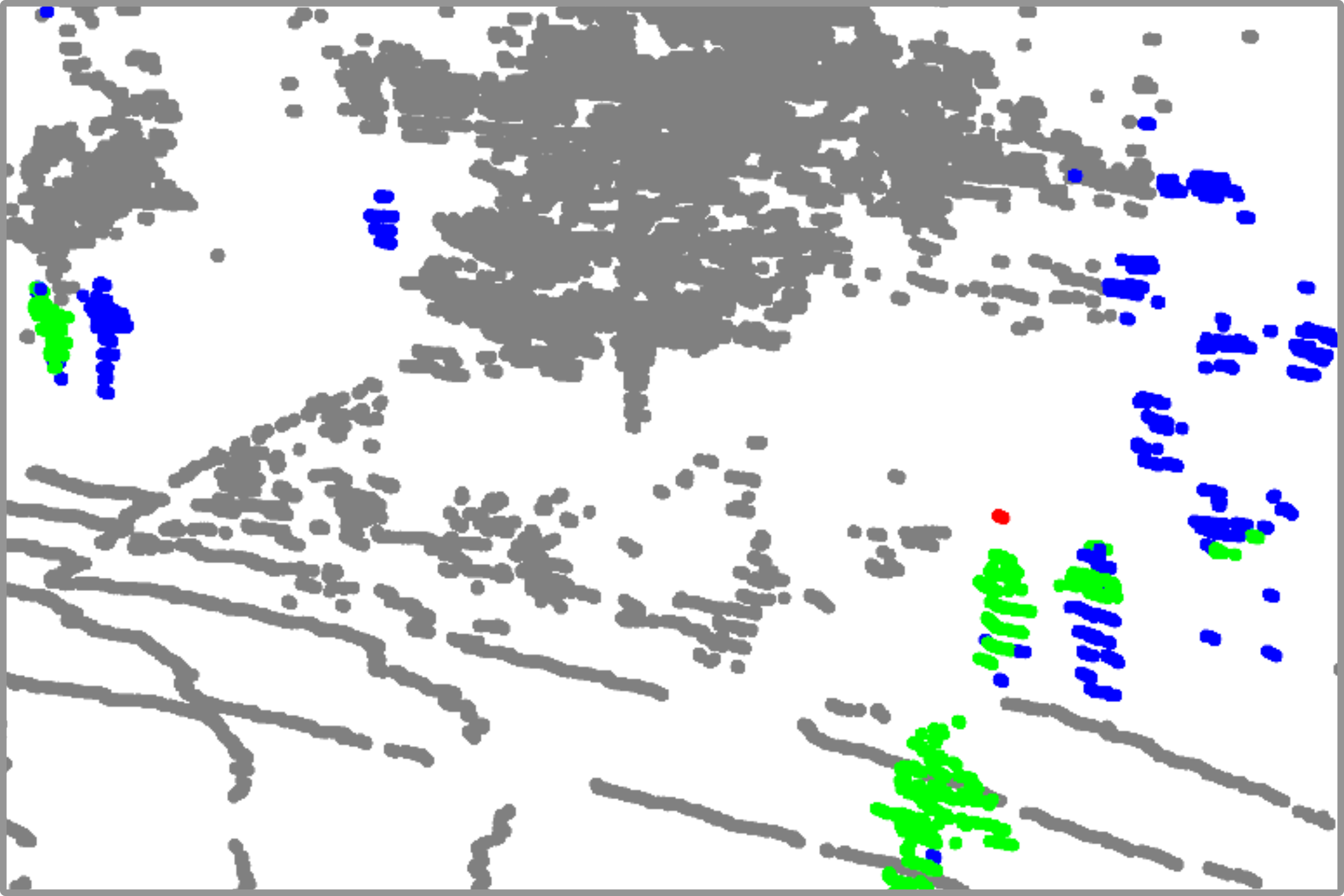}
		\includegraphics[width=0.23\textwidth]{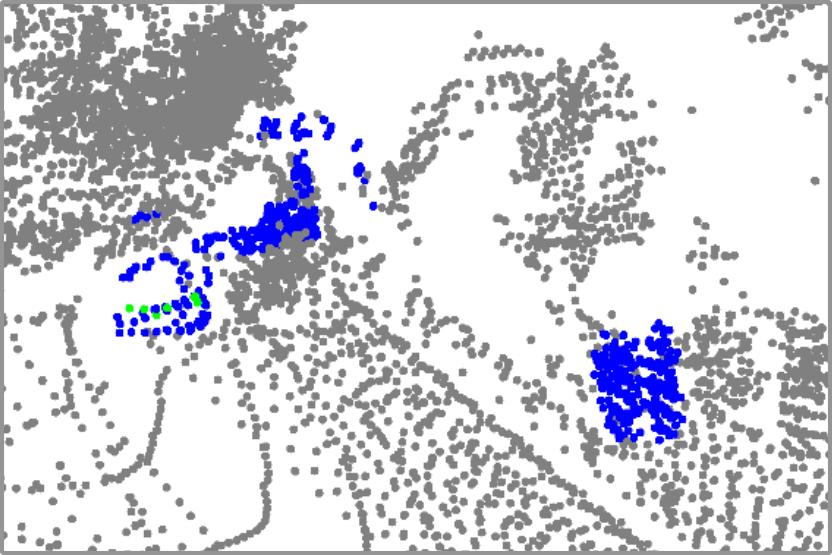}
		\includegraphics[width=0.23\textwidth]{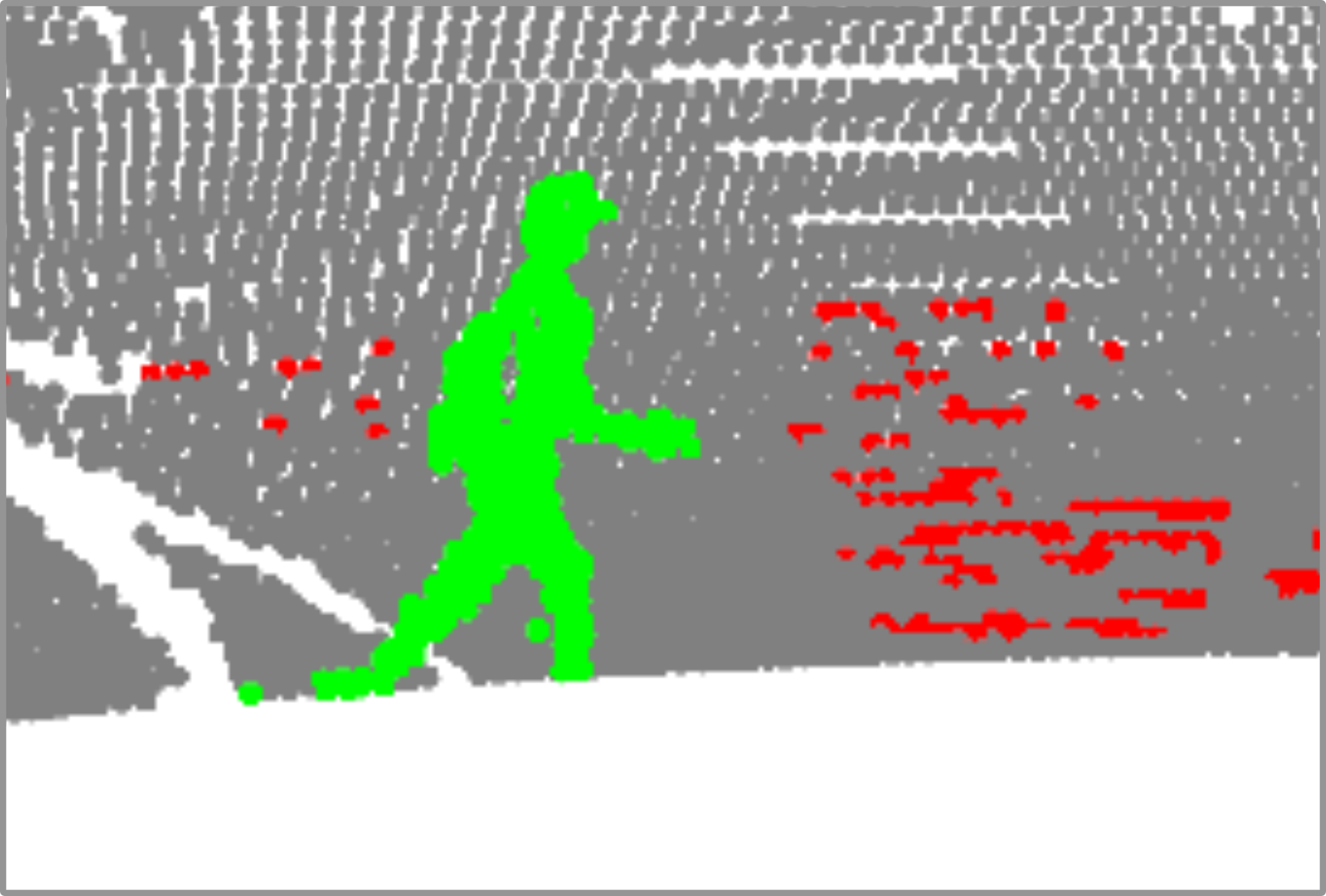}
		\includegraphics[width=0.23\textwidth]{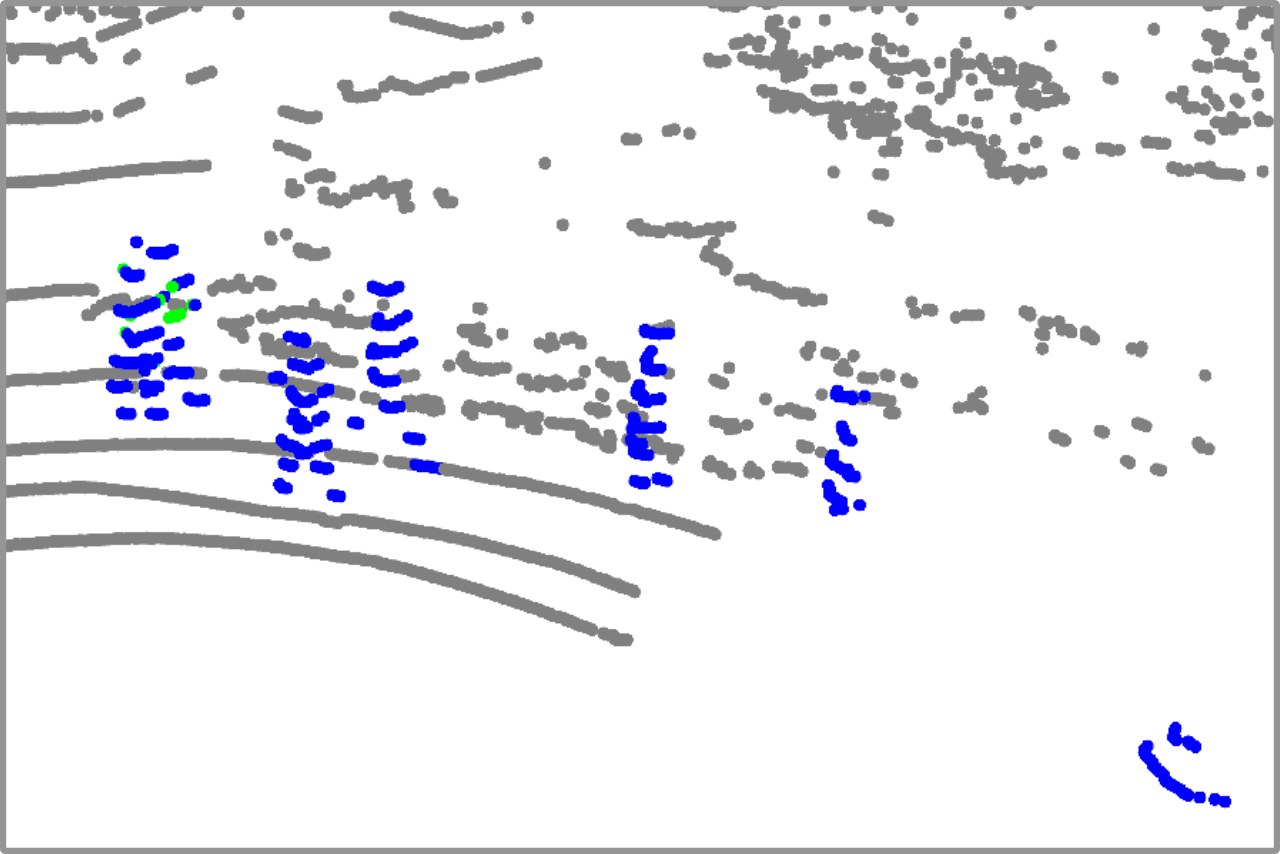}
		\caption{}
	\end{subfigure}
	\begin{subfigure}[b]{0.48\textwidth}
		\centering
		\includegraphics[width=0.23\textwidth]{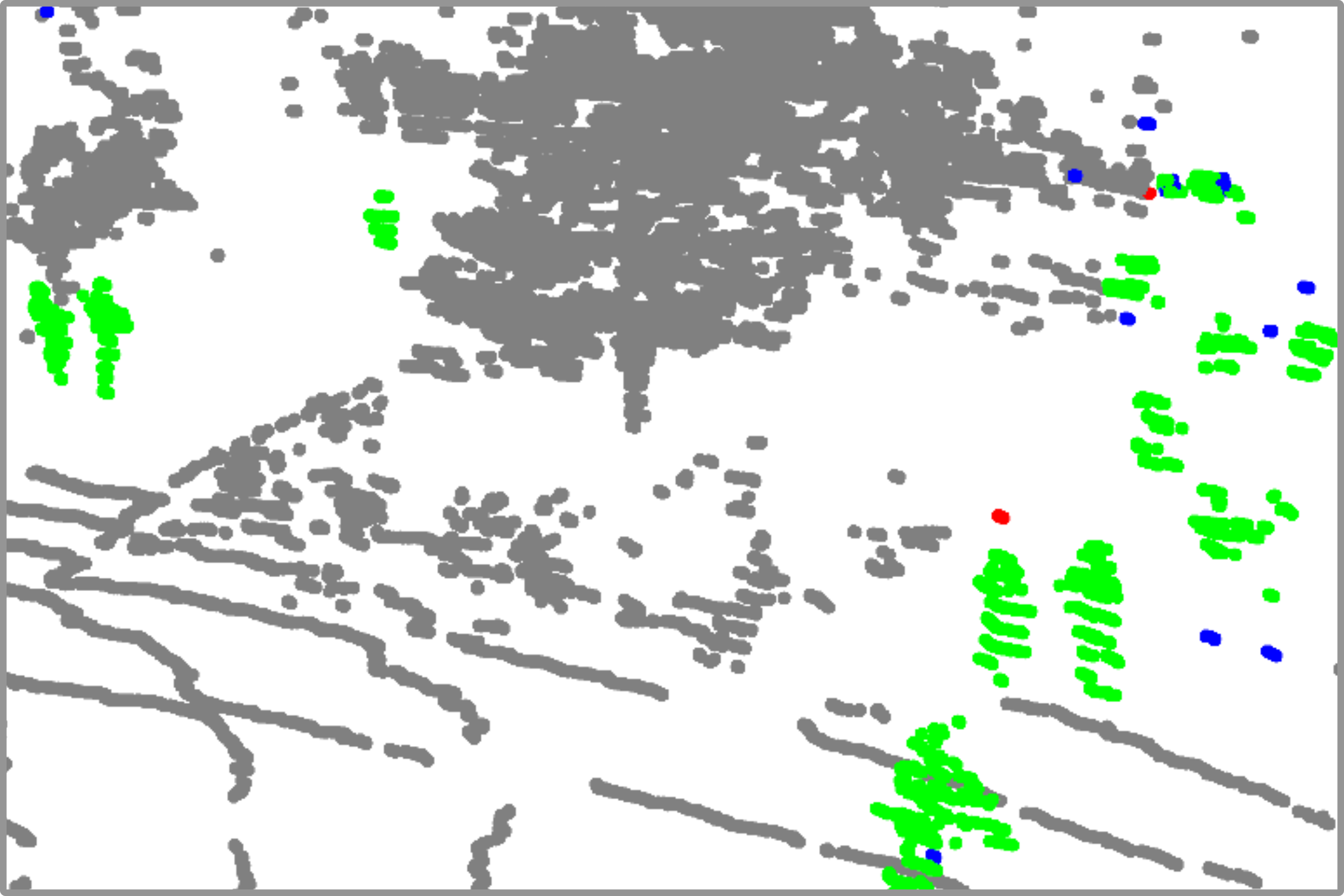}
		\includegraphics[width=0.23\textwidth]{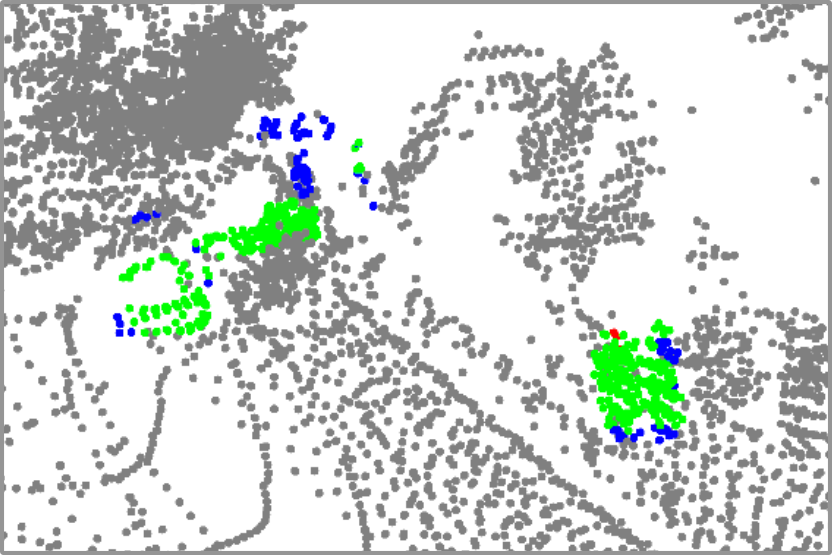}
		\includegraphics[width=0.23\textwidth]{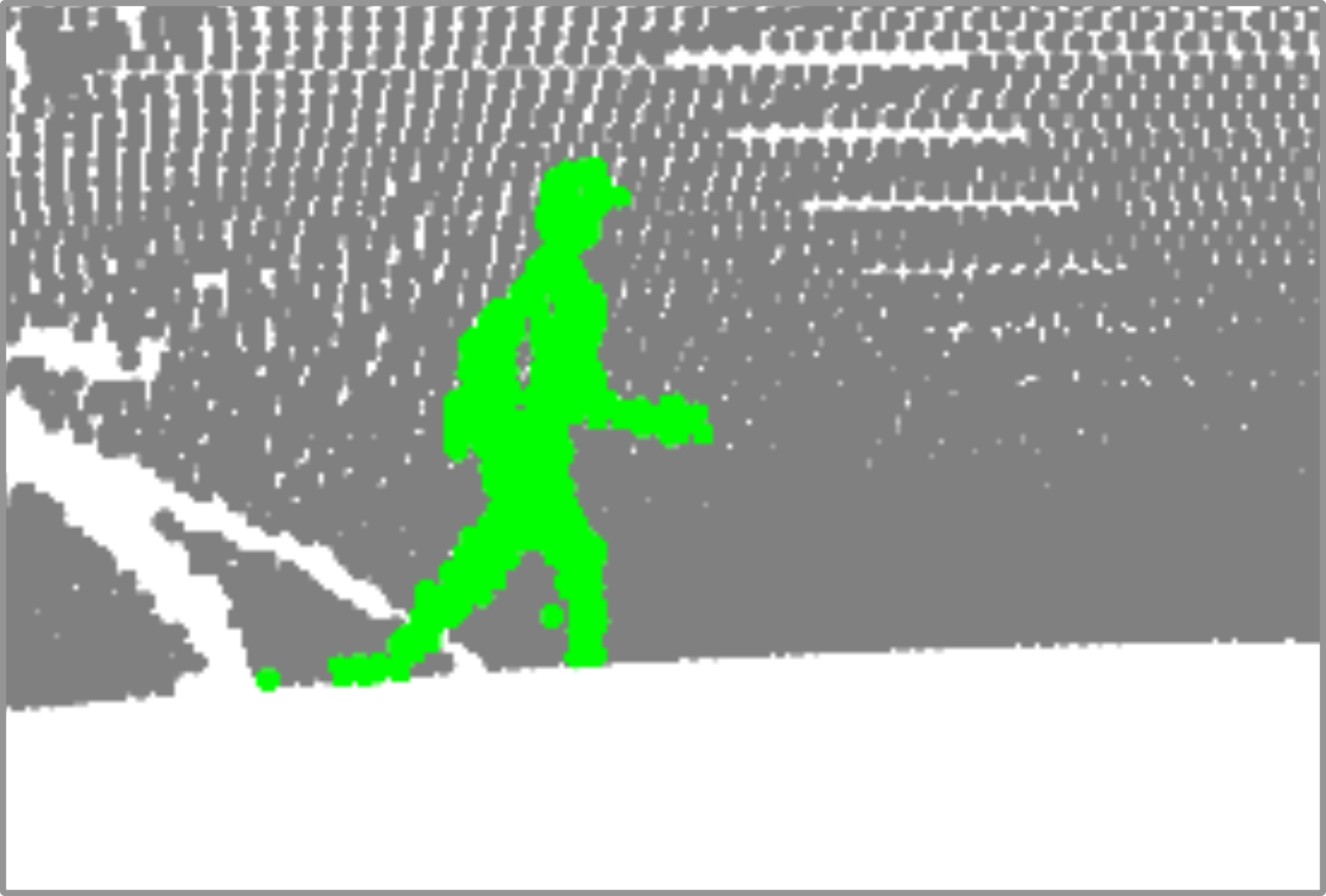}
		\includegraphics[width=0.23\textwidth]{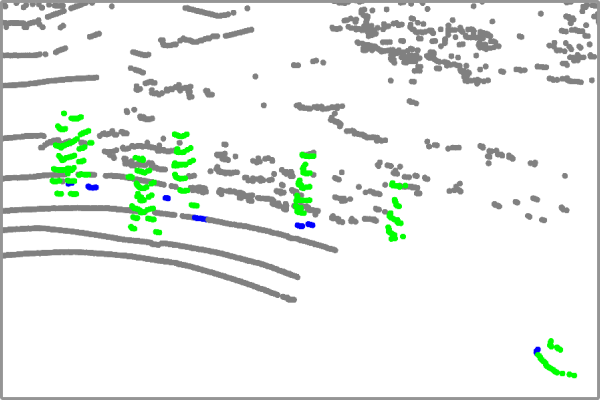}
		\caption{}
	\end{subfigure}
	\caption{Qualitative comparison of MapMOS~\cite{mersch2023building} across all sensors (a)~before and (b)~after training with our dataset. Green, red, and blue points indicate true positives, false positives, and false negatives, respectively. The fewer red and blue points there are, the better~(best viewed in color).}
	\label{fig:before_and_after}
\end{figure}

The next experiment specifically focuses on evaluating the performance changes caused by domain shifts across different LiDAR sensor types within the same environments.
The MOS models showed substantial performance improvements across all sensors after training with our dataset, as presented in \tabref{table:cross_sensor_config} and \figref{fig:before_and_after}.

Interestingly, unlike 4DMOS, whose performance for each test sensor type increased as more diverse training data were provided, the performance of MapMOS showed inconsistency.
MapMOS has better generalization capabilities~\cite{mersch2023building} by taking both a local map and a current scan as inputs.
In particular, the local map is employed to reduce the geometrical differences between each scan from different sensors by accumulating scans over time.
For this reason, even though MapMOS was trained by using \texttt{L}+\texttt{A}, it showed promising performance in~\texttt{O}.
This is because the local maps generated by \texttt{L} and \texttt{A}, and those from \texttt{O} are more similar compared with the raw scans themselves.

Unfortunately, scans from \texttt{V} are too sparse to precisely estimate the relative poses, making local maps sparse and more distorted compared with other sensors.
As a result, MapMOS showed lower IoUs with \texttt{V} in \tabref{table:cross_sensor_config}, meaning relatively poor performance.
Nevertheless, MapMOS was on par with 4DMOS regarding total mean IoU and showed the highest performance in \changed{\texttt{L}, \texttt{V}, and \texttt{O}}.

Therefore, these two experiments imply that there is still room for improvement in making existing MOS methods operate in a sensor-agnostic manner.

\begin{table}[t!]
	\centering
	\captionsetup{font=footnotesize}
	\caption{Mean IoU of MOS approaches when trained solely on data from specific LiDAR sensors.
		The bold texts denote the best performance among all trials and the gray highlights indicate the best results for each method across different training data scenarios~($\texttt{L}$:~Livox Avia, $\texttt{A}$:~Aeva Aeries~\rom{2}, $\texttt{O}$:~Ouster OS2-128, and $\texttt{V}$:~Velodyne VLP-16).}
	\setlength{\tabcolsep}{3.5pt}
	{\scriptsize
		\begin{tabular}{lcccccccc}
			\toprule \midrule
			 \multirow{2}{*}[-0.5em]{Method} & \multirow{2}{*}[-0.5em]{Training data}             & \multicolumn{3}{c}{Solid-state} & \multicolumn{3}{c}{Omnidirectional} & \multirow{2}{*}[-0.5em]{Total}                                                                                         \\ \cmidrule(lr){3-5} \cmidrule(lr){6-8}
			                                               &                                  & \texttt{L}                      & \texttt{A}                          & Avg                            & \texttt{O}          & \texttt{V}          & Avg                 &                     \\ \midrule
\multirow{3}{*}[-0.0em]{4DMOS~\cite{mersch2022ral}} & {\texttt{L}+\texttt{A}} & 72.82                      & 81.58                     & 77.20                     & 72.46               & 47.65          & 70.34          & 68.63               \\
			                                                    & {\texttt{O}+\texttt{V}} & 65.13                      & 74.96                     & 70.05                     & 80.50               & 56.09          & 69.35               & 69.17               \\
			                                                    & {All}                   & \hl{73.72}                 & \hl{84.80}       & \hl{79.26}                & \hl{82.70}               & \hl{\textbf{57.64}} & \hl{\textbf{75.63}} & \hl{\textbf{74.72}} \\ \midrule
			\multirow{3}{*}[-0.0em]{MapMOS~\cite{mersch2023building}} & {\texttt{L}+\texttt{A}}  & \hl{\textbf{75.46}}        & {83.65}                & 79.55       & 77.84          & 25.62               & 68.42               & 65.64          \\
			& {\texttt{O}+\texttt{V}} & 75.00                      & 80.71                     & 77.85                     & \hl{\textbf{87.56}}          & \hl{49.13}               & \hl{74.05}          & \hl{73.10}          \\
			& {All}                   & 75.38                      & \hl{\textbf{85.60}}                     & \hl{\textbf{80.49}}                     & {87.31} & 32.00               & 72.15               & 70.07               \\
			\midrule \bottomrule
		\end{tabular}
	}
	\label{table:cross_sensor_config}
\end{table}

\subsection{Automatic Labeling Performance}\label{sec:static_map_building}

Finally, we demonstrate the superiority of our automatic labeling framework.
Because AutoMOS uses ERASOR, which is outperformed by ERASOR2 (see \tabref{table:static_map_building_in_synced_scans}), a direct comparison would be unfair.
Therefore, we separately evaluate the performance of (a) static map building approaches for initial MOS labeling and (b) tracking-based filtering.

\begin{figure}[t]
    \captionsetup{font=footnotesize}
    \centering
    \begin{subfigure}[b]{0.13\textwidth}
        \centering
        \includegraphics[width=1.0\textwidth]{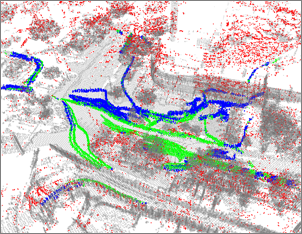}
        \includegraphics[width=1.0\textwidth]{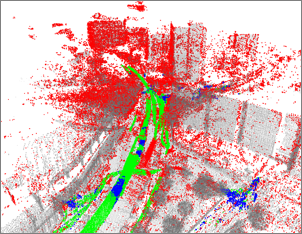}
        \includegraphics[width=1.0\textwidth]{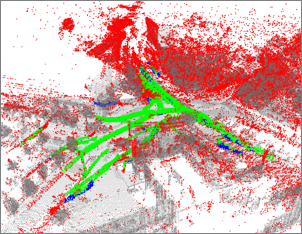}
        \caption{Removert~\cite{kim2020iros}}
    \end{subfigure}
    \begin{subfigure}[b]{0.13\textwidth}
        \centering
        \includegraphics[width=1.0\textwidth]{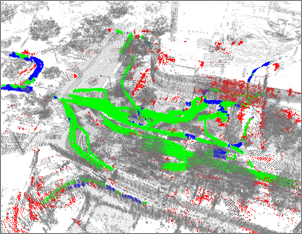}
        \includegraphics[width=1.0\textwidth]{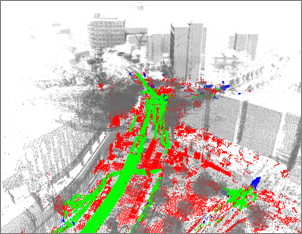}
        \includegraphics[width=1.0\textwidth]{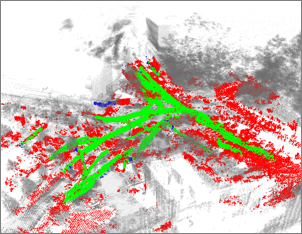}
        \caption{ERASOR~\cite{lim2021ral}}
    \end{subfigure}
    \begin{subfigure}[b]{0.13\textwidth}
        \centering
        \includegraphics[width=1.0\textwidth]{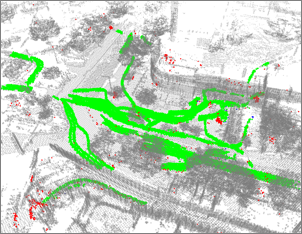}
        \includegraphics[width=1.0\textwidth]{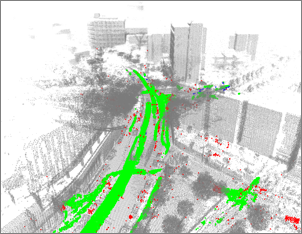}
        \includegraphics[width=1.0\textwidth]{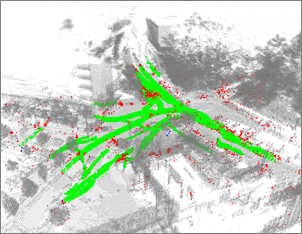}
        \caption{ERASOR2~\cite{lim2023erasor2}}
    \end{subfigure}
    \vsfigu
    \caption{(a)-(c) Qualitative comparison of static map building results produced by state-of-the-art methods on our dataset using synced scans.
    Green, red, and blue points indicate true positives, false positives, and false negatives, respectively. The fewer red and blue points there are, the better~(best viewed in color).}
    \label{fig:sota_static_map_building_comparison}
    \vsfig
\end{figure}

First, to more explicitly show the performance differences in static map building, we conducted experiments on the three most crowded scenes.
As a result, we demonstrate that ERASOR2 shows a substantially higher $\text{F}_1$ score compared with Removert, a range image-based approach, and ERASOR, an initial MOS labeling module in the Auto-MOS, as shown in \tabref{table:static_map_building_in_synced_scans}.
In particular, ERASOR showed lower PR and RR than ERASOR2.
This is because ERASOR directly subtracts the estimated dynamic points from the map cloud without considering instance information, incorrectly estimating static points as dynamic while leaving some dynamic points on the map, as described in \figref{fig:sota_static_map_building_comparison}(b).
In contrast, by leveraging instance information, ERASOR2 precisely rejected traces of moving objects in the map cloud while preserving most static points, as presented in \figref{fig:sota_static_map_building_comparison}(c).
This implies that ERASOR2 consistently labels the dynamic points within each scan.

\newcommand{\threemetrics}{\multicolumn{1}{c}{\begin{tabular}[c]{@{}c@{}} $\PRmap$ [\%] \end{tabular}}  & \multicolumn{1}{c}{\begin{tabular}[c]{@{}c@{}} $\RRmap$ [\%] \end{tabular}} & $\text{F}_{1}$ score}

\begin{table}[t!]
	\centering
	\captionsetup{font=footnotesize}
	\caption{Comparison of static map building approaches for the most crowded frame sequences in our dataset~($\PRmap$: Preservation Rate, $\RRmap$: Rejection Rate).}
{\scriptsize
	\begin{tabular}{clccc}
		\toprule \midrule
		Frame range & Method & \threemetrics \\ \midrule
		\multirow{3}{*}{2,250-2,500}
& Removert~\cite{kim2020iros}   & 85.072 & 47.170 & 0.607    \\
		& ERASOR~\cite{lim2021ral}      & 95.325 & 82.490 & 0.884    \\
		& ERASOR2~\cite{lim2023erasor2} & \textbf{99.522} & \textbf{95.339} & \textbf{0.974} \\ \midrule
		\multirow{3}{*}{8,600-8,800}
& Removert~\cite{kim2020iros}   & 80.581 & 71.965 &	0.760    \\
		& ERASOR~\cite{lim2021ral}      & 91.610 & 84.290 & 0.878    \\
		& ERASOR2~\cite{lim2023erasor2} & \textbf{99.530} & \textbf{93.740} & \textbf{0.965}    \\ \midrule
		\multirow{3}{*}{11,070-11,300}
& Removert~\cite{kim2020iros}   & 82.852 & 81.301 & 0.821    \\
		& ERASOR~\cite{lim2021ral}      & 93.969 & 89.955 & 0.919    \\
		& ERASOR2~\cite{lim2023erasor2} & \textbf{99.676} & \textbf{97.175} & \textbf{0.984}  \\ \midrule
		\bottomrule
	\end{tabular}
	}
	\label{table:static_map_building_in_synced_scans}
\end{table}
 
\begin{table}[t!]
	\centering
	\captionsetup{font=footnotesize}
	\caption{Mean IoU before and after the application of tracking-based filtering approaches.}
{\scriptsize
	\begin{tabular}{lcccc}
		\toprule \midrule
		Method & $\iou$ \\ \midrule
ERASOR2~\cite{lim2023erasor2}   &  21.8 \\
		ERASOR2 + Tracking-based filtering in Auto-MOS~\cite{chen2022ral} & 53.4     \\
		ERASOR2 + Our tracking-based filtering & \textbf{61.2} \\ \midrule
		\bottomrule
	\end{tabular}
	}
	\label{table:comparison_of_tracking}
\end{table}

Second, as shown in \tabref{table:comparison_of_tracking}, the tracking-based filtering in Auto-MOS showed a substantial performance increase, which indicates that it significantly reduces the number of false positive points.
However, as described in \secref{sec:tracking_based_filtering}, it cannot reduce the number of false negatives.
In contrast, by introducing augmented bounding boxes, our approach could suppress the impact of false negative points.
By doing so, our proposed filtering showed higher $\iou$.

Therefore, we conclude that the combination of ERASOR2 and our proposed tracking-based filtering is a suitable automatic labeling framework to help human labelers reduce the time needed for manual labeling.

\section{Conclusion}
\label{sec:conclusion}

In this paper, we have presented a novel moving object segmentation dataset for heterogeneous LiDAR sensors.
Furthermore, we have proposed a novel instance-aware automatic labeling framework to reduce the time cost and effort of a human labeler when annotating labels in large-scale scenes.
Finally, we demonstrated the necessity of a heterogeneous LiDAR moving object segmentation dataset by suggesting new research directions towards sensor-agnostic segmentation, enabling better evaluations in this field of research.

In future work, we will further study domain generalization of MOS approaches in terms of different environments and different settings between existing datasets and our HeLiMOS.

\vspace{-0.20cm}
\section*{Acknowledgements}

Above all things, we thank Prof. Ayoung Kim's group, particularly Minwoo Jung, for making their HeLiPR dataset~\cite{jung2023helipr} and development tools available.
\vspace{-0.10cm}

\bibliographystyle{URL-IEEEtrans}

\bibliography{main.bib}

\end{document}